\documentclass[table,xcdraw]{article} 
\usepackage{iclr2023_conference,times}

\usepackage{amsmath,amsfonts,bm}

\def\eqref#1{equation~\ref{#1}}

\def\1{\bm{1}}

\def\rb{{\textnormal{b}}}

\def\vb{{\bm{b}}}

\DeclareMathAlphabet{\mathsfit}{\encodingdefault}{\sfdefault}{m}{sl}
\SetMathAlphabet{\mathsfit}{bold}{\encodingdefault}{\sfdefault}{bx}{n}

\DeclareMathOperator*{\argmin}{arg\,min}

\let\ab\allowbreak

\usepackage{hyperref}
\usepackage{url}

\newcommand{\norm}[1]{\left\|#1\right\|}

\usepackage[vlined,ruled]{algorithm2e}

\usepackage[utf8]{inputenc} 
\usepackage[T1]{fontenc}    
\usepackage{hyperref}       
\usepackage{url}            
\usepackage{booktabs}       
\usepackage{amsfonts}       
\usepackage{nicefrac}       
\usepackage{microtype}      
\usepackage{thm-restate}

\usepackage{microtype}
\usepackage{graphicx}
\usepackage[FIGTOPCAP]{subfigure}
\usepackage{changes}

\usepackage{amsmath}
\usepackage{multirow}
\usepackage{hhline}
\usepackage{wrapfig}
\usepackage{floatrow}
\usepackage{caption}
\usepackage{enumitem}
\usepackage[export]{adjustbox}
\usepackage{amsmath,mathtools}

\usepackage{tikz}

\usepackage{algorithmic}

\makeatletter
\newcommand{\raisemath}[1]{\mathpalette{\raisem@th{#1}}}
\newcommand{\raisem@th}[3]{\raisebox{#1}{$#2#3$}}
\makeatother

\newcommand{\kals}{{\texttt{Kaleidoscopic}~}}

\definecolor{armygreen}{rgb}{0.29, 0.33, 0.13}
\definecolor{bittersweet}{rgb}{1.0, 0.44, 0.37}
\definecolor{aoenglish}{rgb}{0.0, 0.5, 0.0}

\title{Generative Kaleidoscopic Networks}

\author{
  ~~~~~\quad\qquad Harsh Shrivastava
  \hspace{0mm}\\
  \begin{tabular}{c}
      $\prescript{}{}{\text{Microsoft Research, Redmond, USA}}$\\
      {\tiny Contact: hshrivastava@microsoft.com}
  \end{tabular}
}

\iclrfinalcopy 
\begin{document}

\maketitle

\begin{abstract}
We discovered that the neural networks, especially the deep ReLU networks, demonstrate an `over-generalization' phenomenon. That is, the output values for the inputs that were not seen during training are mapped close to the output range that were observed during the learning process. In other words, the neural networks learn a many-to-one mapping and this effect is more prominent as we increase the number of layers or depth of the neural network. We utilize this property of neural networks to design a dataset kaleidoscope, termed as `Generative Kaleidoscopic Networks'. Briefly, if we learn a model to map from input $x\in\mathbb{R}^D$ to itself $f_\mathcal{N}(x)\rightarrow x$, the proposed \kals sampling procedure starts with a random input noise $z\in\mathbb{R}^D$ and recursively applies $f_\mathcal{N}(\cdots f_\mathcal{N}(z)\cdots )$. After a burn-in period duration, we start observing samples from the input distribution and the quality of samples recovered improves as we increase the depth of the model. \textit{Scope}: We observed this phenomenon to various degrees for the other deep learning architectures like CNNs, Transformers \& U-Nets and we are currently investigating them further.

%
\textit{Software \& Demo}: {\tiny \url{https://github.com/Harshs27/generative-kaleidoscopic-networks}}
\end{abstract}

\section{Introduction}

We started this work as an exercise to understand whether there is a way for neural networks to learn fractals. Fractals are interesting structures that are often generated by repeatedly applying a set of (simplish) rules over a seemingly modest starting point. Since, fractals are widely observed in nature, we thought that it would be a great exercise to approximate or reverse-engineer a given fractal. We chose neural networks for this task, due to their expressive power in terms of complex function representations. Now, that is an ambitious task! Anyway, we started with exploring properties of neural networks to see if we can replicate a `Kaleidoscope' type of behaviour, where we can keep switching from one data point (eg. image) to another. During this process, we stumbled upon an interesting phenomenon about neural networks and that is the topic of this paper. We further utilized this property and designed a sampling algorithm. Since, we can handle multi-modal data, one can potentially use this technique for doing Generative AI based tasks. We now describe our findings.




\begin{figure}[b]
\centering 
\includegraphics[width=135mm]{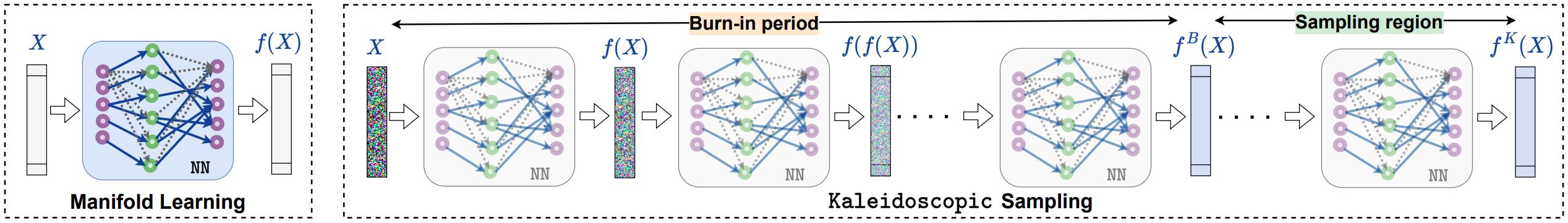}
\caption{\small \textbf{Manifold learning \& \kals sampling}. [left] During the manifold learning process, the deep ReLU networks or the  Multilayer Perceptron (MLP) weights have their gradients enabled, shaded in {\color{blue}blue}, and their output units bounded between $(0,1)$ via a `Sigmoid' non-linearity (or between $(-1,1)$ for `Tanh'). They are learned as per Eq.~\ref{eq:manifold-learning}. [right] During the sampling process, the weights of the neural network model are frozen and the input is a randomly initialized noise, sampled from a normal or uniform distribution. The model $f$ is repeatedly applied to the input noise in accordance to Eq.~\ref{eq:manifold-sampling}. Once the function is applied `B' number of times $f\circ f\circ \cdots \circ f$, we start obtaining the samples closer to the input data distribution $B\rightarrow K$.}
\label{fig:manifold-learning-sampling}
\end{figure}

\section{Understanding the manifolds learned by neural networks}

We start by studying the manifolds learned by Neural Networks (NNs) by analysing their function behaviour especially for the input ranges that are not observed during the training process. We chose an architecture that maps data from one manifold to another and study the following observation

~\fbox{%
    \parbox{0.98\textwidth}{%
        \textit{\textbf{Over-generalization phenomenon}: If the output units of neural networks are bounded, they tend to `over-generalize' on the input data. That is, the output values over the entire input range are mapped close to the output range that were seen during learning, exhibiting a `many-to-one' mapping.}
    }%
}



\begin{figure}
\begin{center}
\subfigure[]{\includegraphics[width=0.25\textwidth]{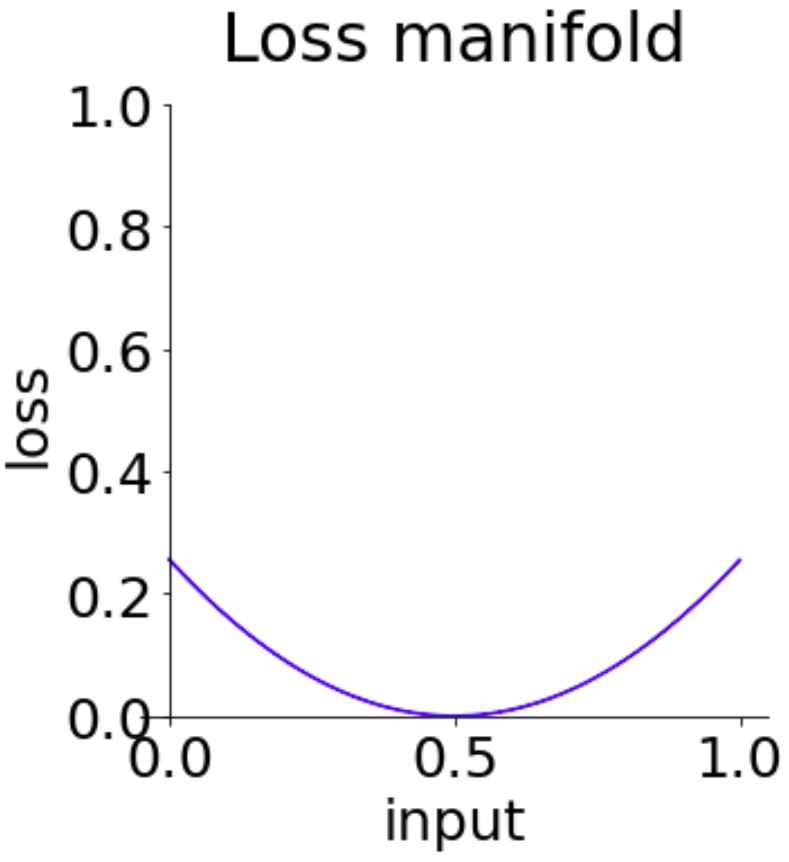}~}\quad
\subfigure[]{\includegraphics[width=0.5\textwidth]{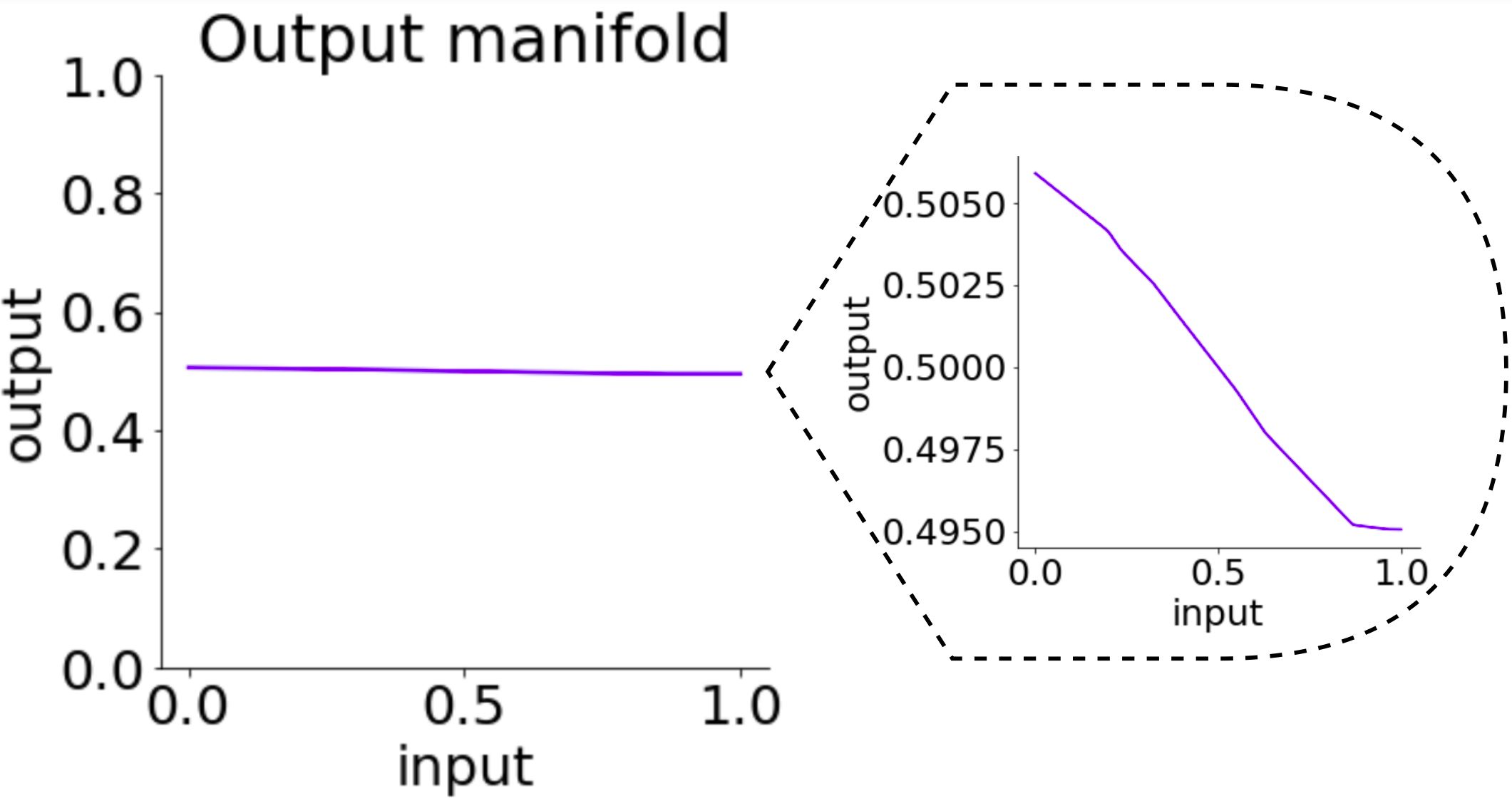}~}
\subfigure[]{\includegraphics[width=0.25\textwidth]{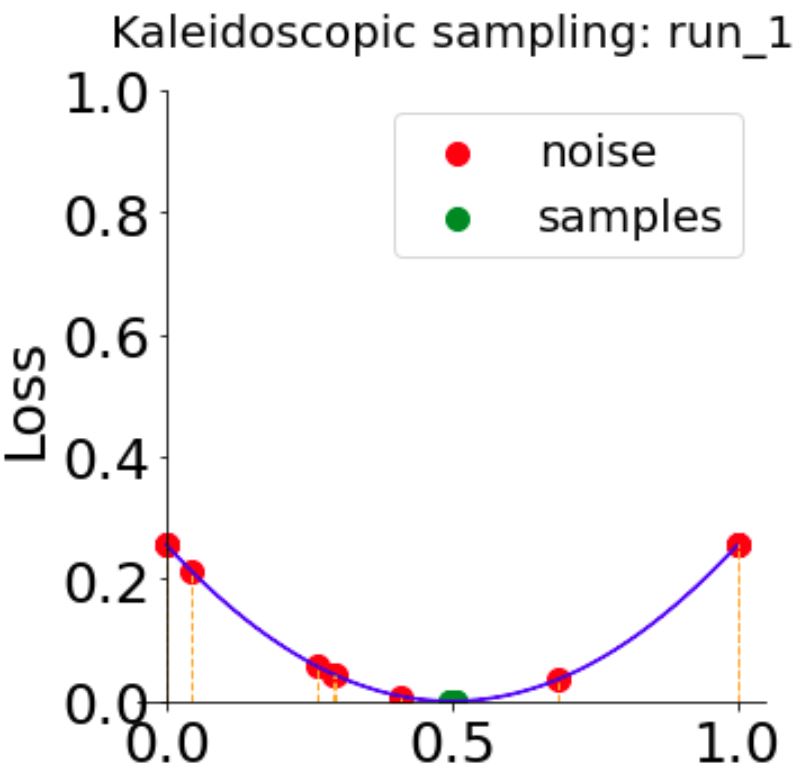}~}
\subfigure[]{\includegraphics[width=0.25\textwidth]{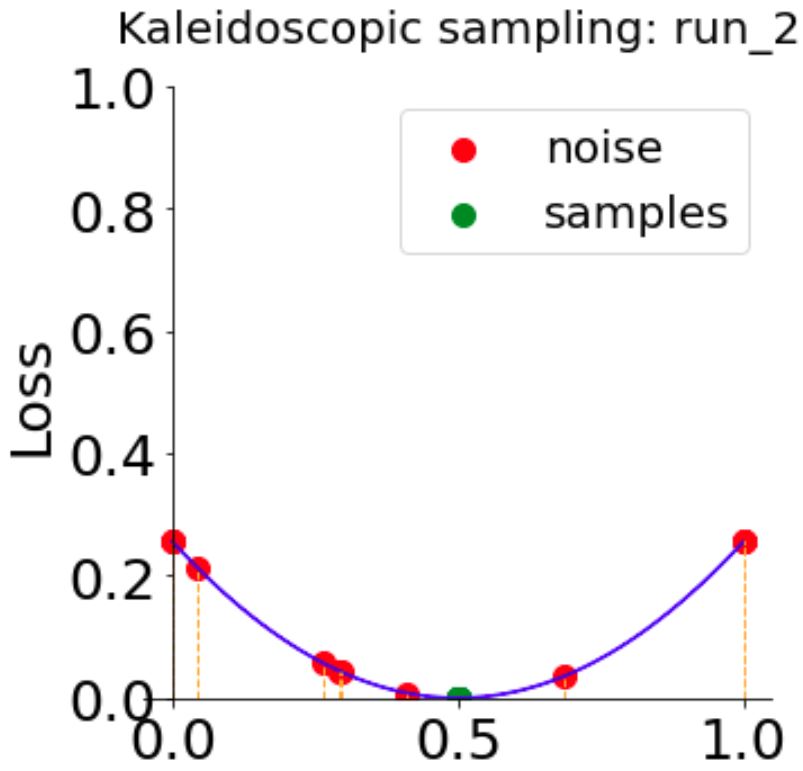}~}
\subfigure[]{\includegraphics[width=0.25\textwidth]{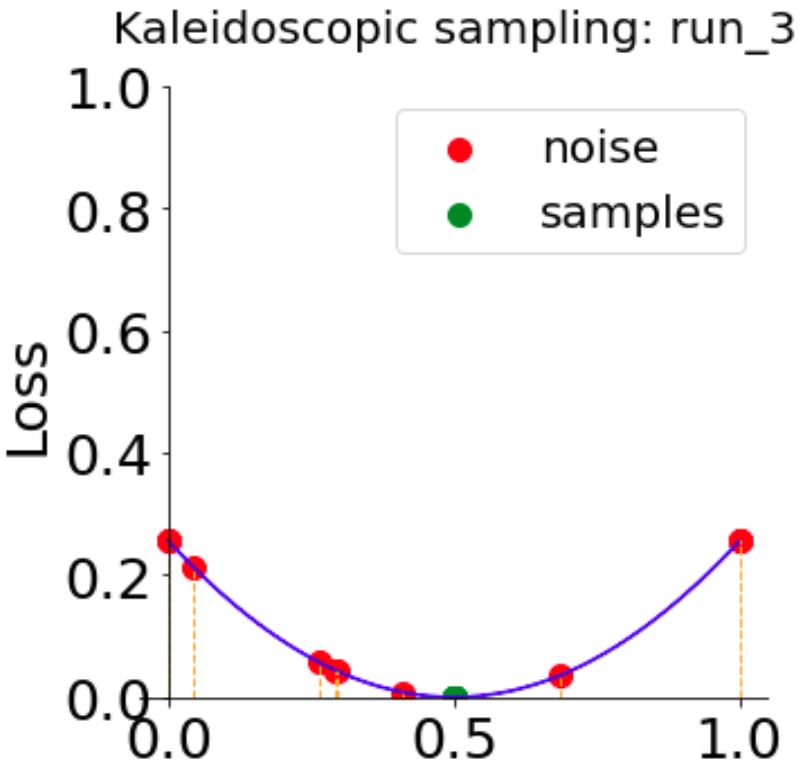}}
\end{center}
\caption{\small \textbf{Single point in 1D space}. The top row shows the \textit{over-generalization} phenomenon by performing manifold learning on a single point $x=0.5$, where range of $x\in(0, 1)$. A MLP with number of hidden layer $L=2$, hidden units size $H=5$  with `ReLU' as the non-linearity in the middle layers and an entrywise `Sigmoid' in the final layer was trained till the loss on the input data point was very low ($\leq 1e{-8}$, epochs $\geq 1K$). In other words, we overfit the MLP to the input data (we observed similar loss profiles with other choices of $L$ and $H$). (a) We plot the loss function values for the entire range of input. As expected, the value at $x=0.5$ is the lowest. (b) This is particularly interesting to observe that the output values observed of the MLP for the entire range of input values, is close to $0.5$ (as it matches input $x=0.5$ to same value at output). This close-up part on the right also shows that the neural network tend to learn \textit{many-to-one} mapping. The bottom row shows the \kals sampling results for the MLP model. The initial noise (10 points) are shown in {\color{bittersweet} red} and the recovered samples are shown in {\color{aoenglish}green}. (c-e) The model is progressively applied $1\rightarrow 3$ times and the samples obtained are close to the training input, as expected.(best viewed in color)}
\label{fig:single-point-1D}
\end{figure}

\textbf{Manifold Learning}: We choose the class of deep learning models that maps (or transforms) data from one manifold to another, which are architectures like the Autoencoder~\cite{goodfellow2016deep} or a Neural Graphical Model~\cite{shrivastava2023neural}. We chose a Multi-layer Perceptron (MLP) with either `Sigmoid' or `Tanh' function at the output layer. Given a set $X\in\mathbb{R}^{M\times D}$ with M samples of D dimensions each, we consider a deep learning architecture $f_\mathcal{N}$ that operates on $X\in\mathbb{R}^{M\times D}$ and minimizes the following loss function 
\begin{align}\label{eq:manifold-learning}
    (\mathcal{L}_{\text{manifold}})~~~~\min_{f_\mathcal{N}} \frac{1}{M\cdot D}\sum_i^M\norm{X^{(i)}-f_{\mathcal{N}}(X^{(i)})}_2^2
\end{align}
We divide by $D$, so that the maximum loss value is $1$, since the non-linearity applied at the output bounds the range. Fig.~\ref{fig:manifold-learning-sampling} [left] shows the learning procedure for the manifold mapping architecture which reconstructs the input at the model's output. Our learning setup is flexible to the choice of the deep learning architecture which can change based on the input data to get a good fit for the above loss function. We can also view this manifold learning approach as learning a probability density function over the input data, $p(X) \propto e^{-\norm{X - f_\mathcal{N}(X)}_2^2}$.
Now, we will understand how well the MLPs fit a function to an input data and their manifold profile based on optimizing Eq.\ref{eq:manifold-learning} loss.

\begin{figure}
\centering 
($H=5, L=2$)
\subfigure[]{\includegraphics[width=0.5\textwidth]{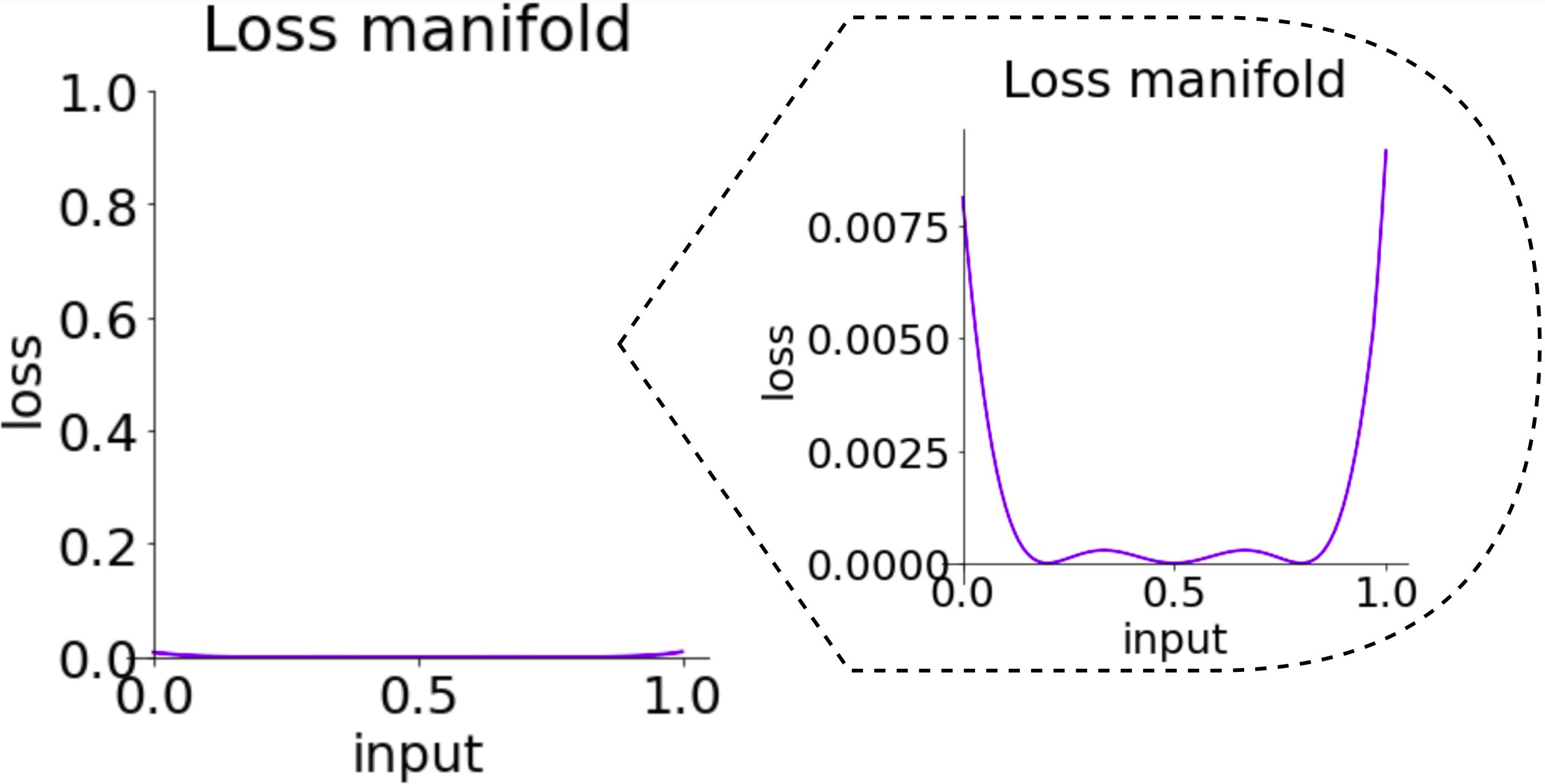}~}\quad
\subfigure[]{\includegraphics[width=0.25\textwidth]{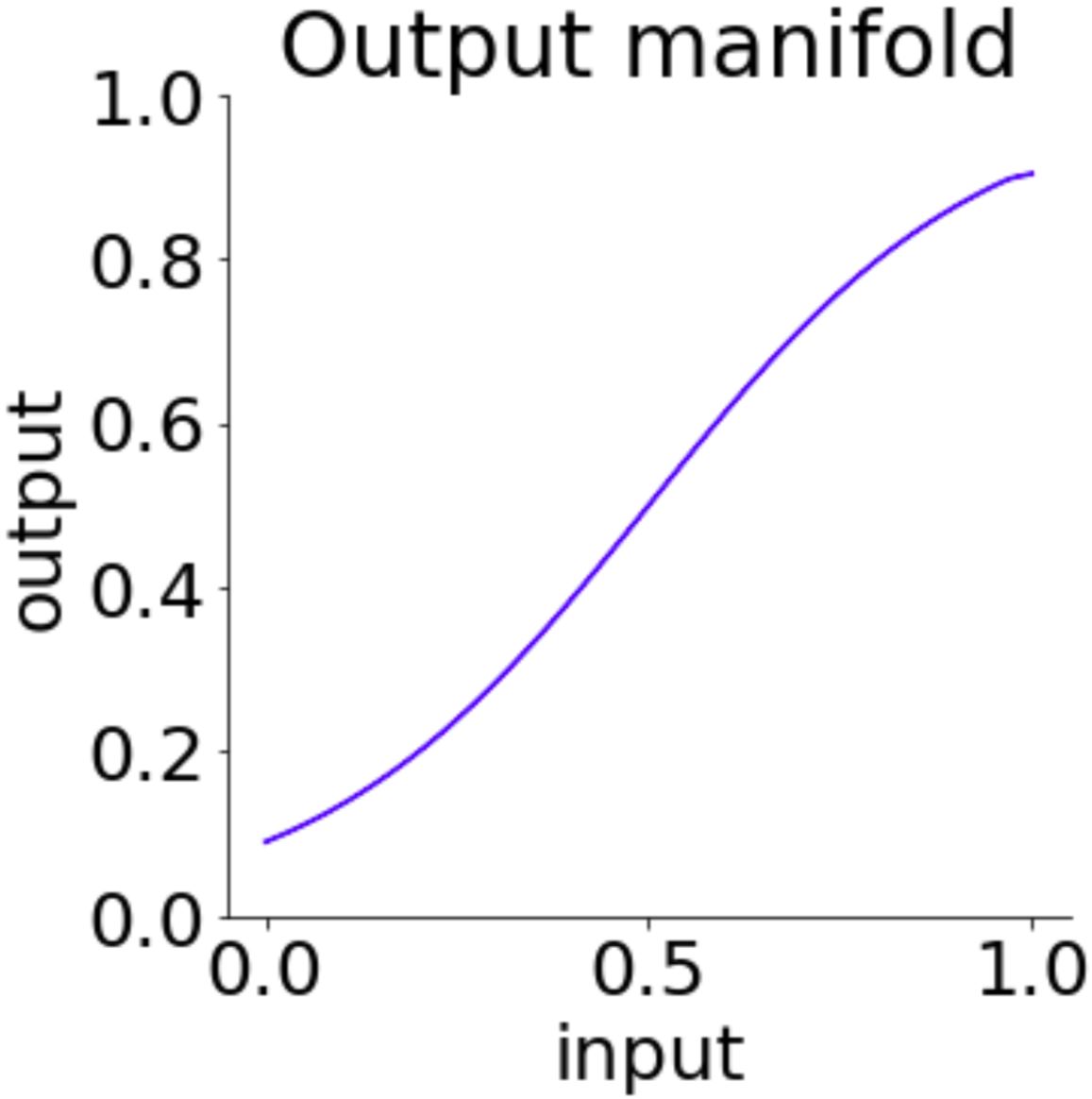}~}
\subfigure[]{\includegraphics[width=0.24\textwidth]{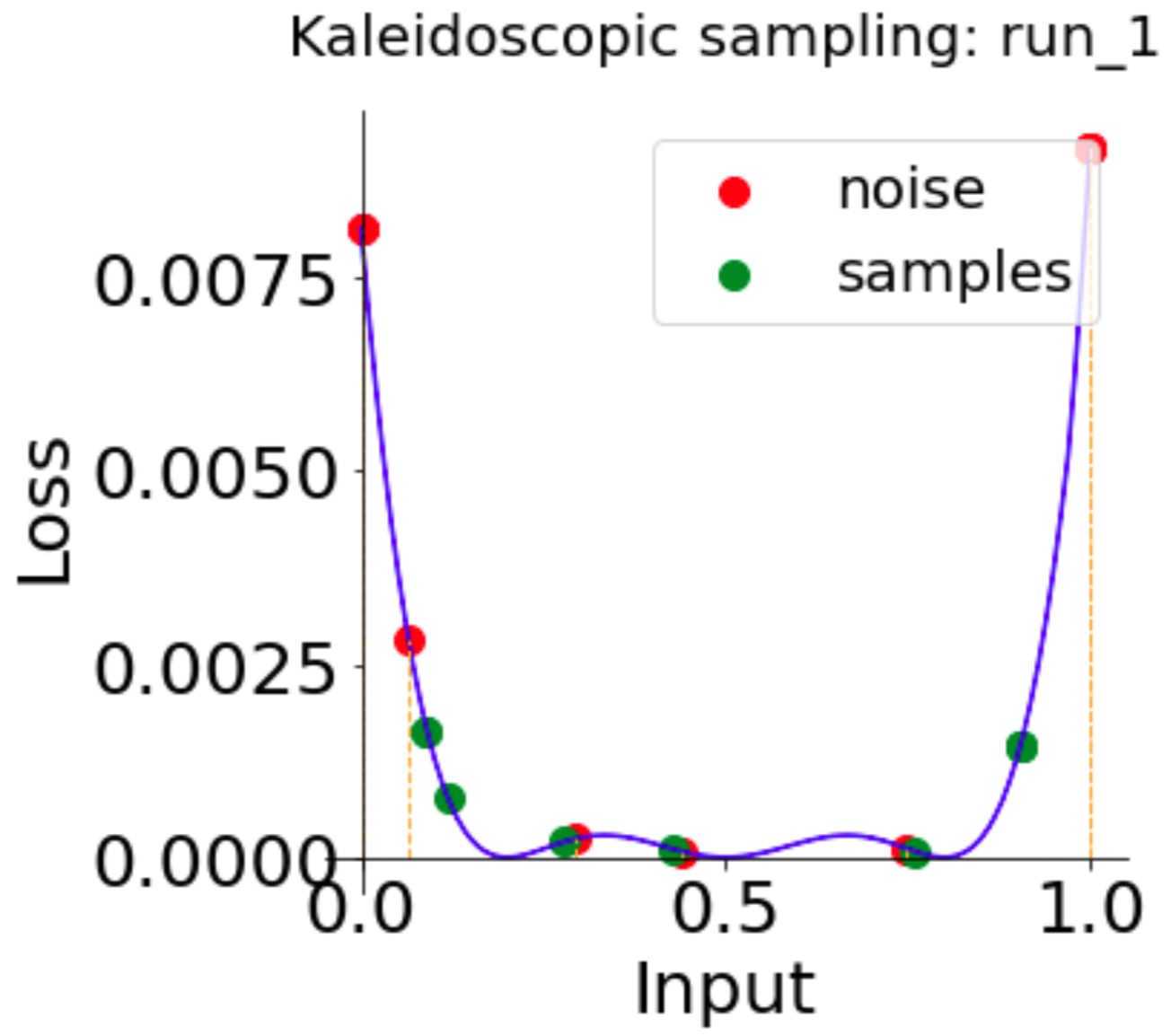}~}
\subfigure[]{\includegraphics[width=0.24\textwidth]{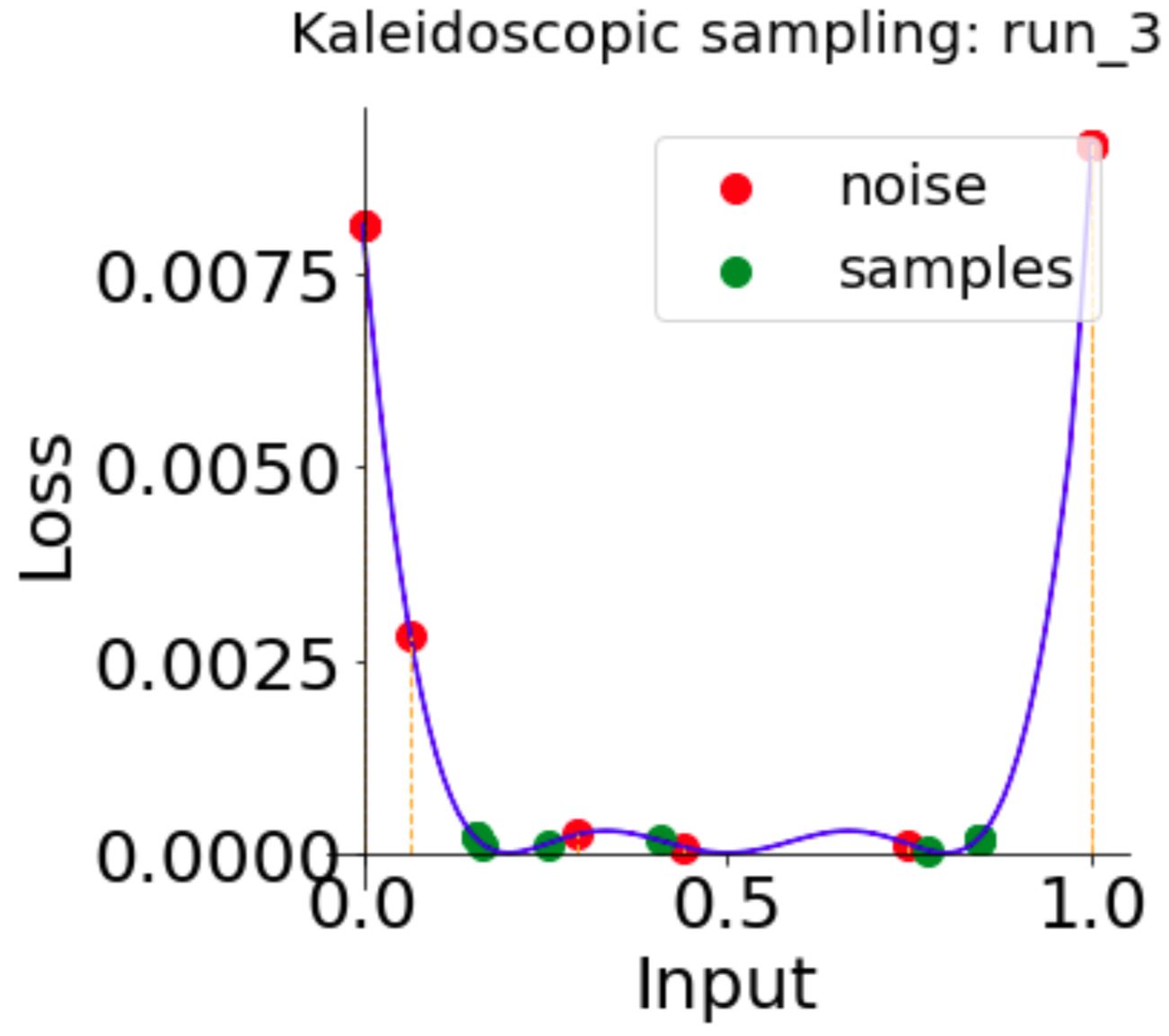}~}
\subfigure[]{\includegraphics[width=0.24\textwidth]{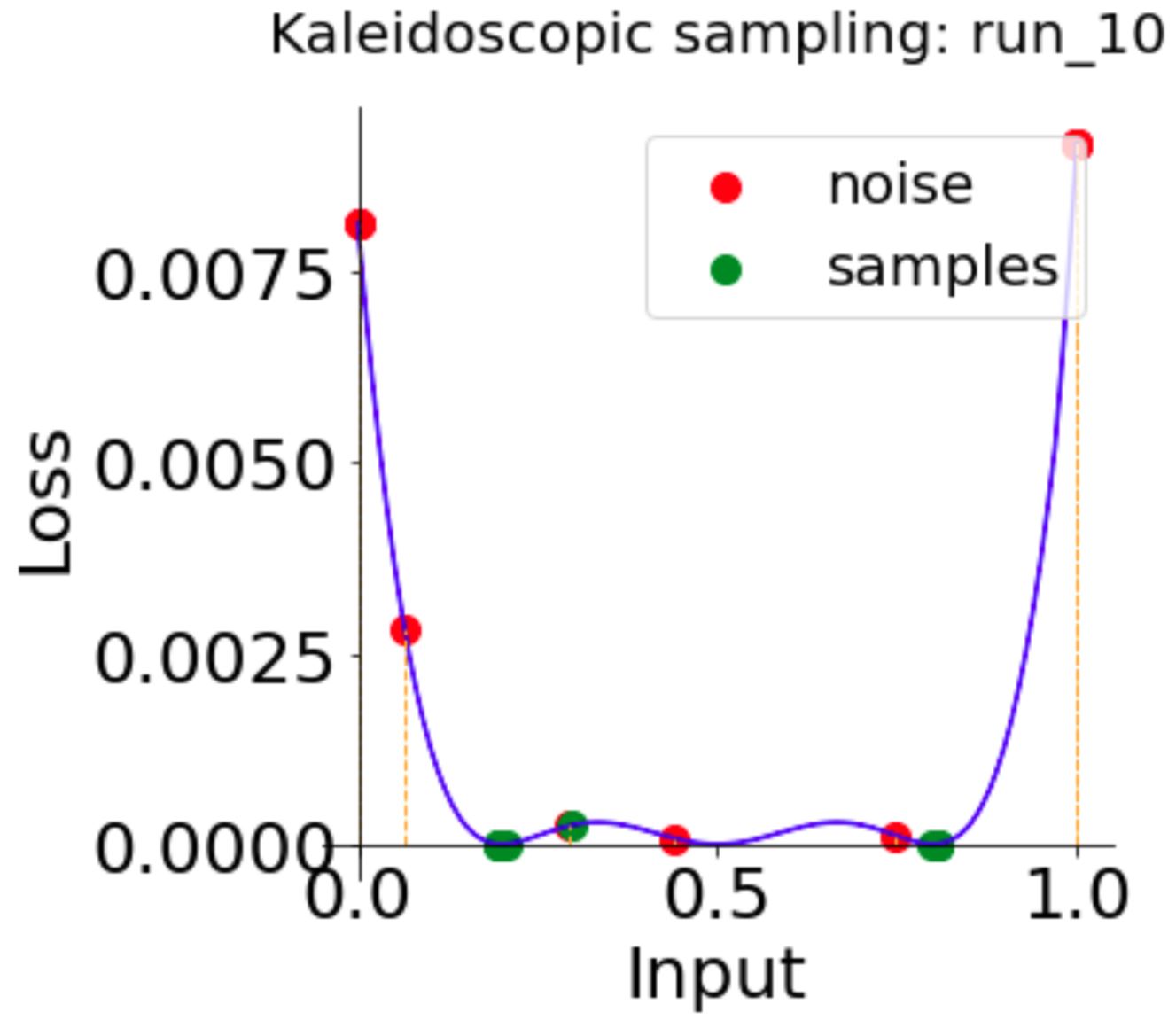}}
\subfigure[]{\includegraphics[width=0.24\textwidth]{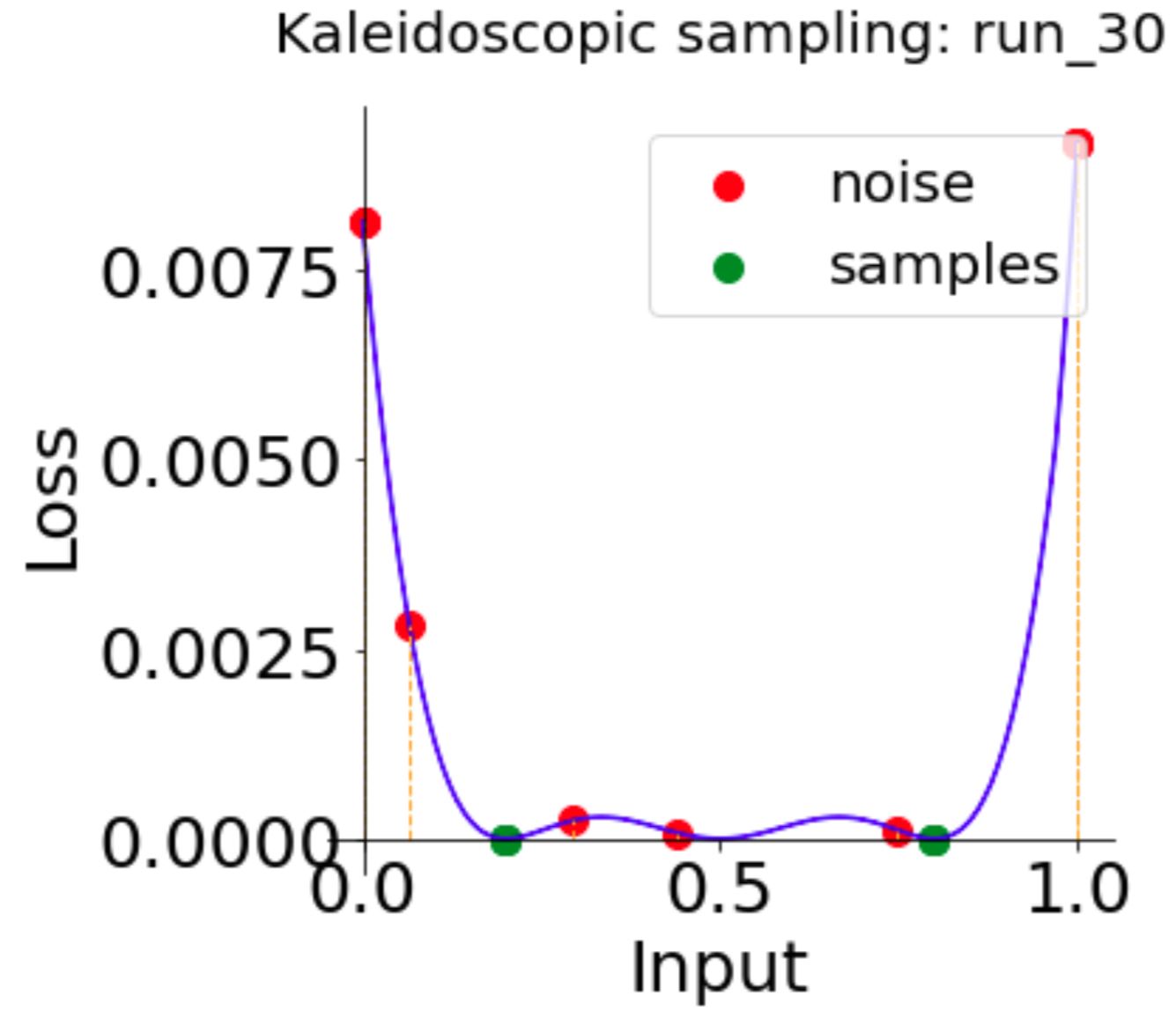}}
\rule{\textwidth}{1pt}
($H=5, L=7$)
\subfigure[]{\includegraphics[width=0.5\textwidth]{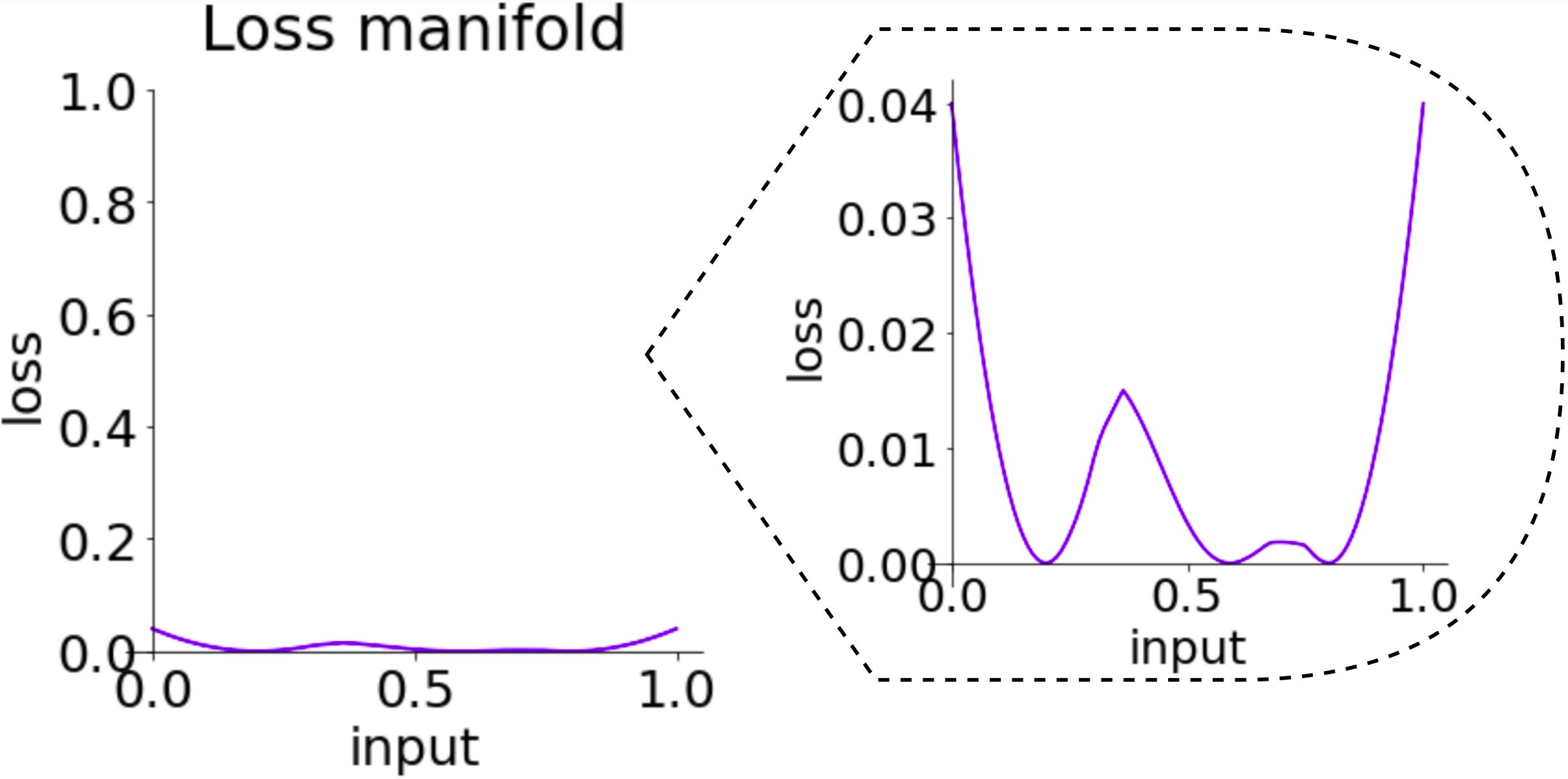}~}\quad
\subfigure[]{\includegraphics[width=0.25\textwidth]{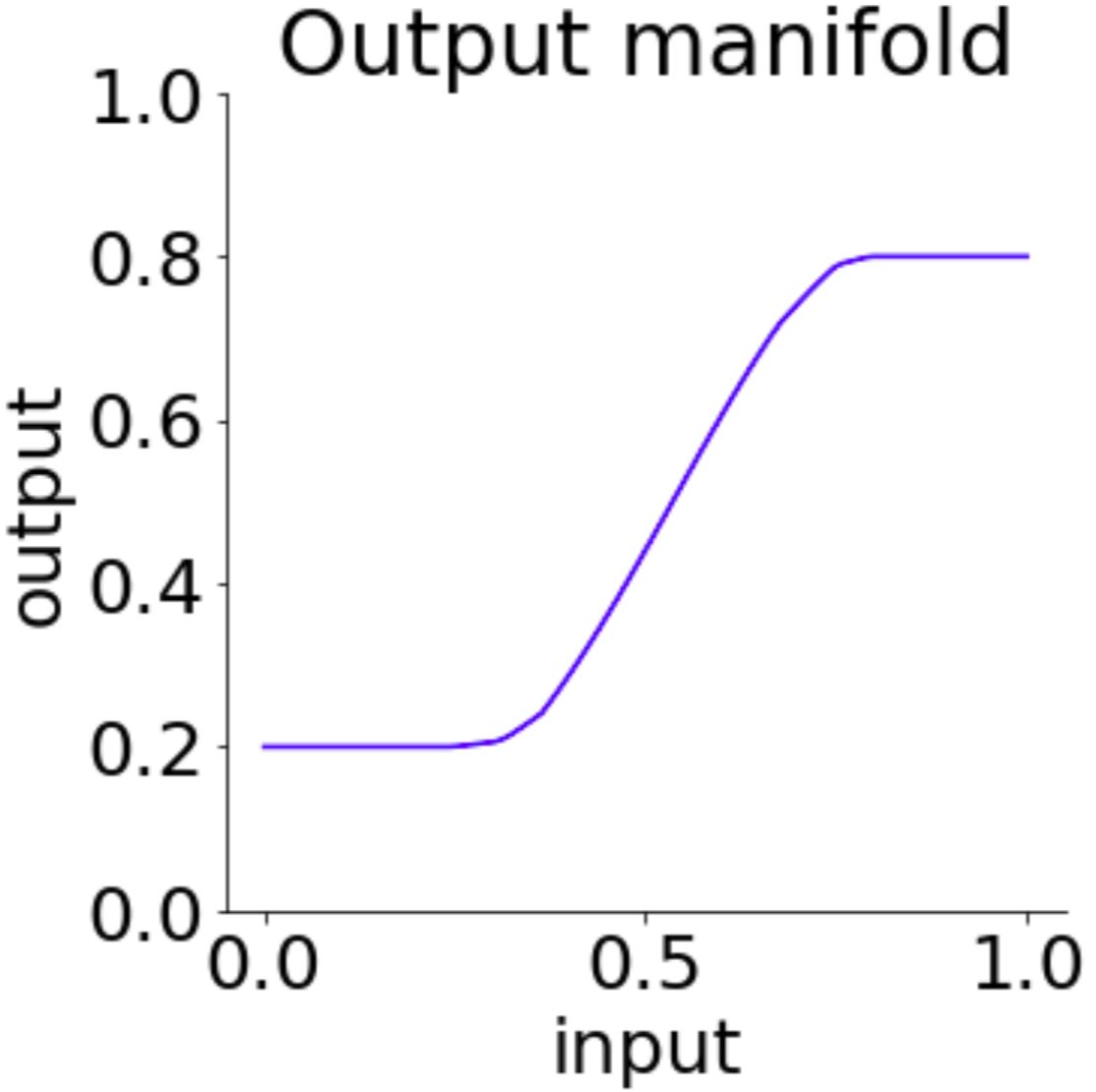}~}
\subfigure[]{\includegraphics[width=0.24\textwidth]{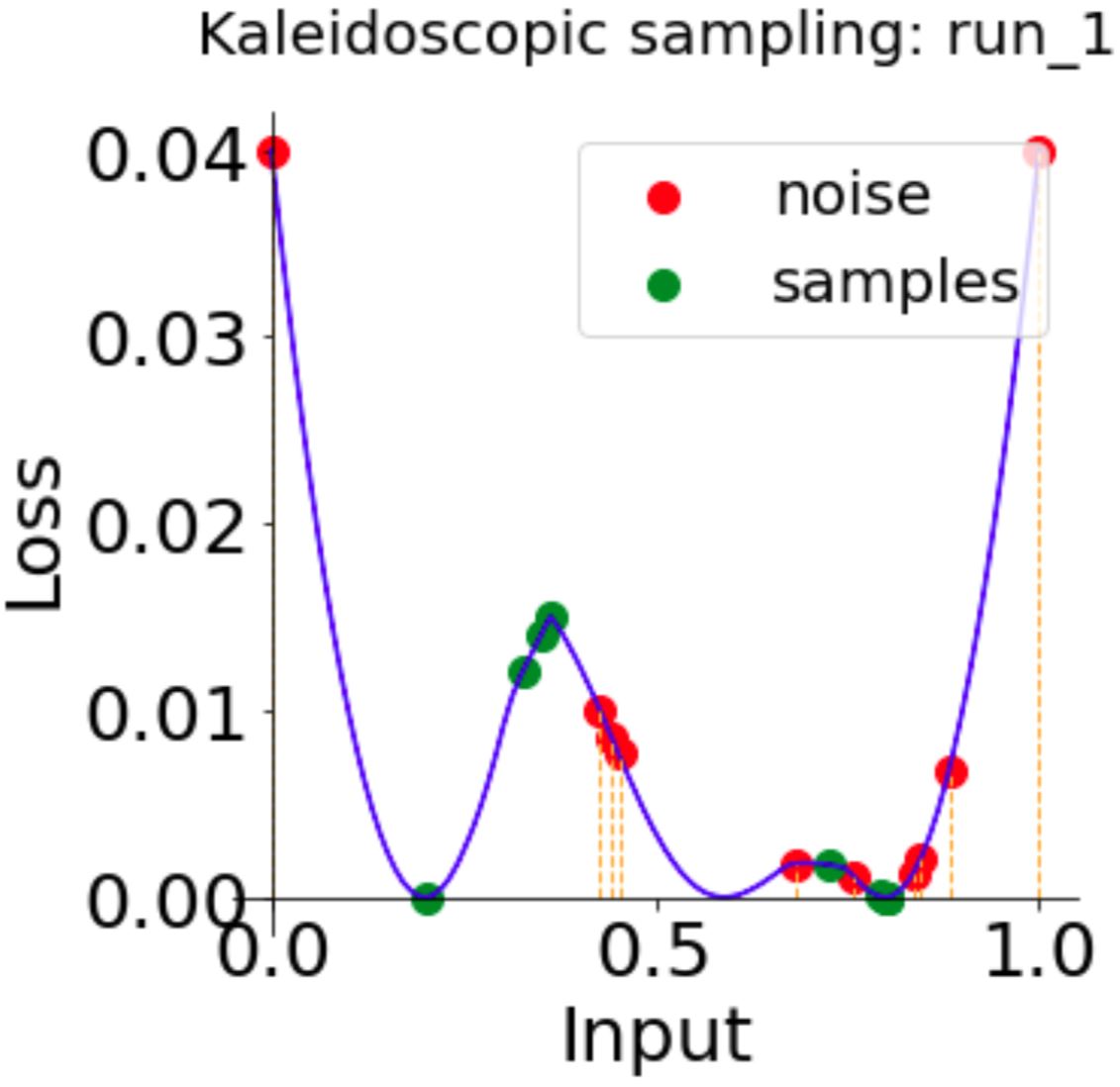}~}
\subfigure[]{\includegraphics[width=0.24\textwidth]{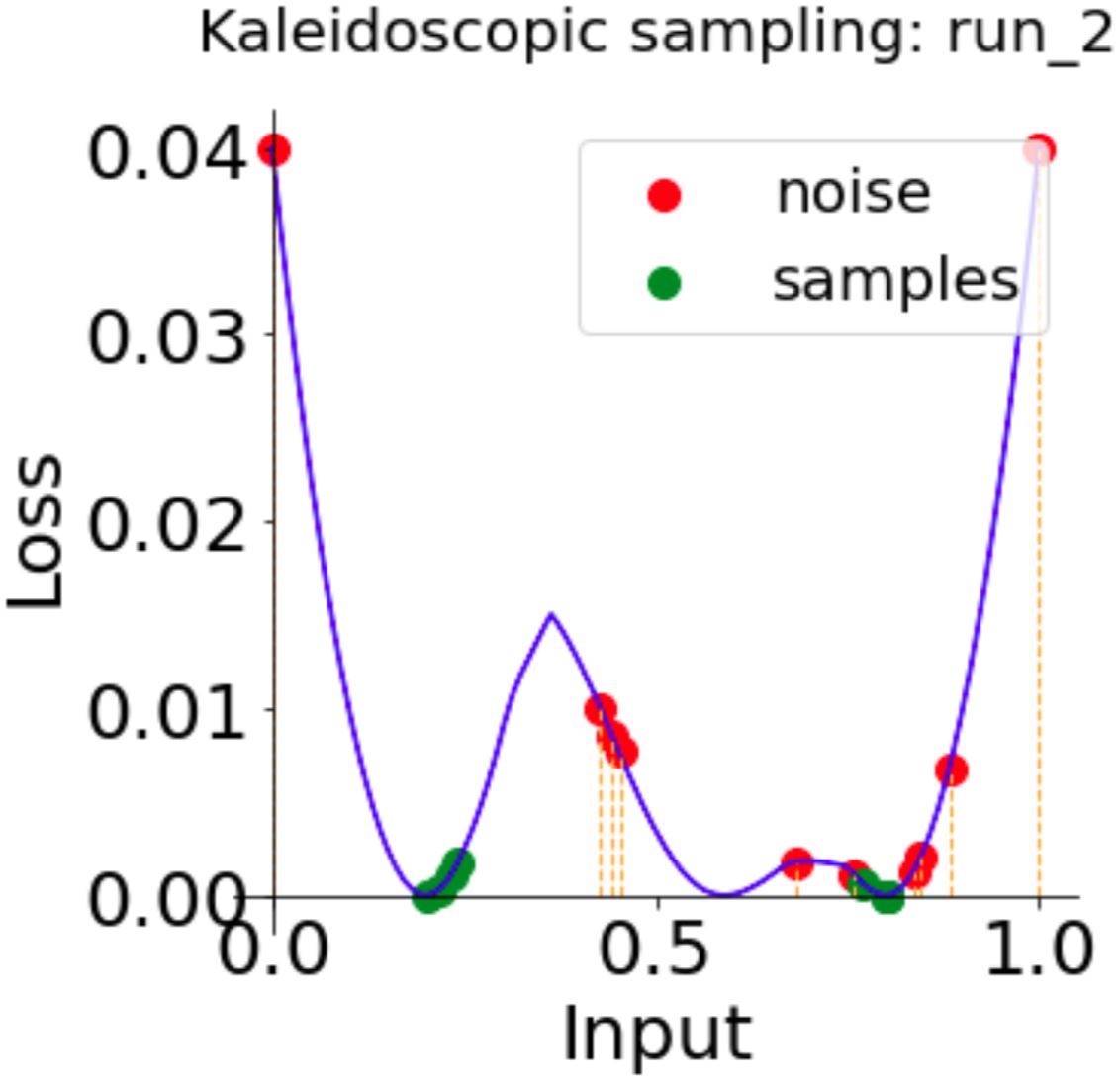}}
\subfigure[]{\includegraphics[width=0.24\textwidth]{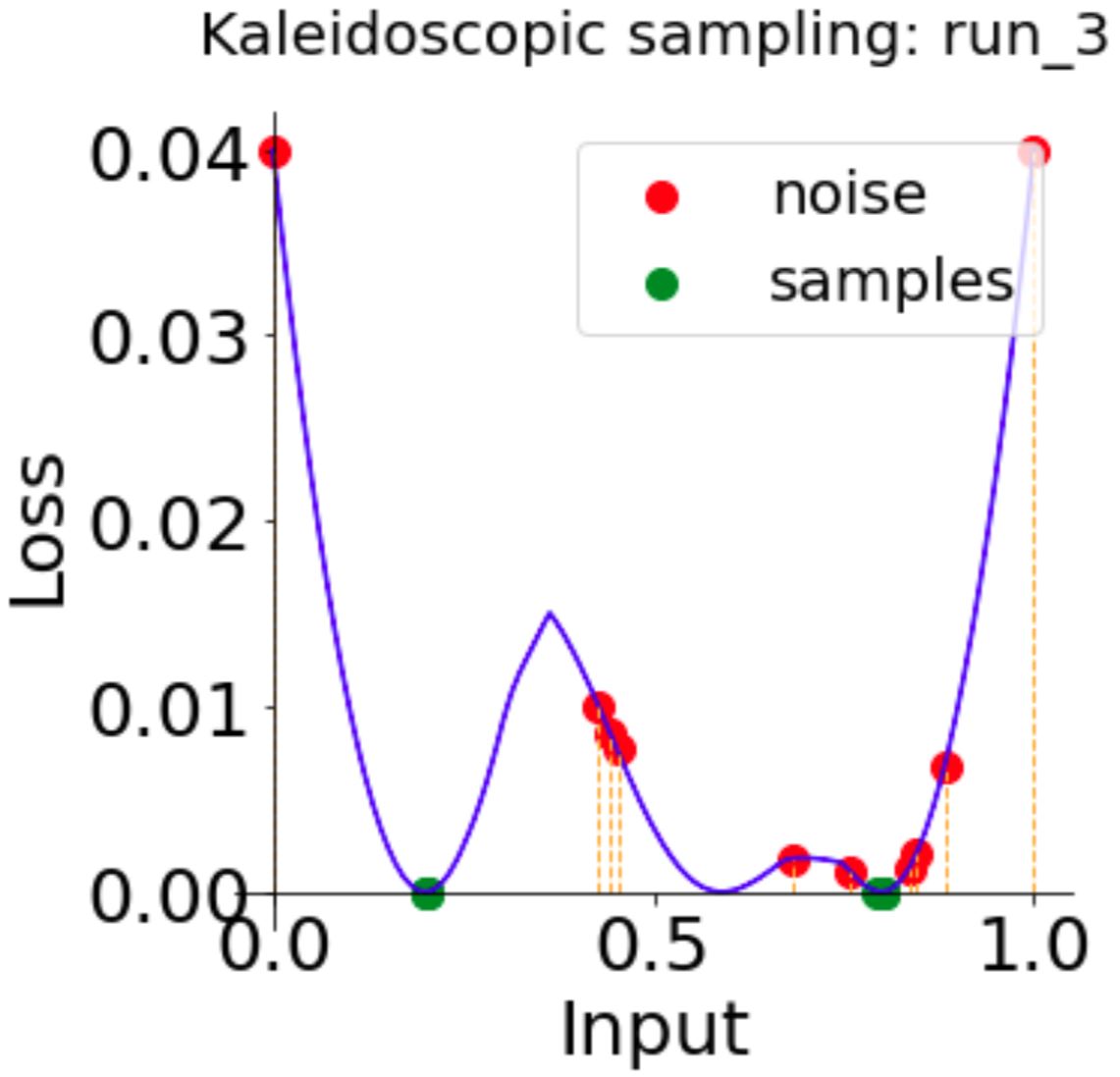}}
\subfigure[]{\includegraphics[width=0.24\textwidth]{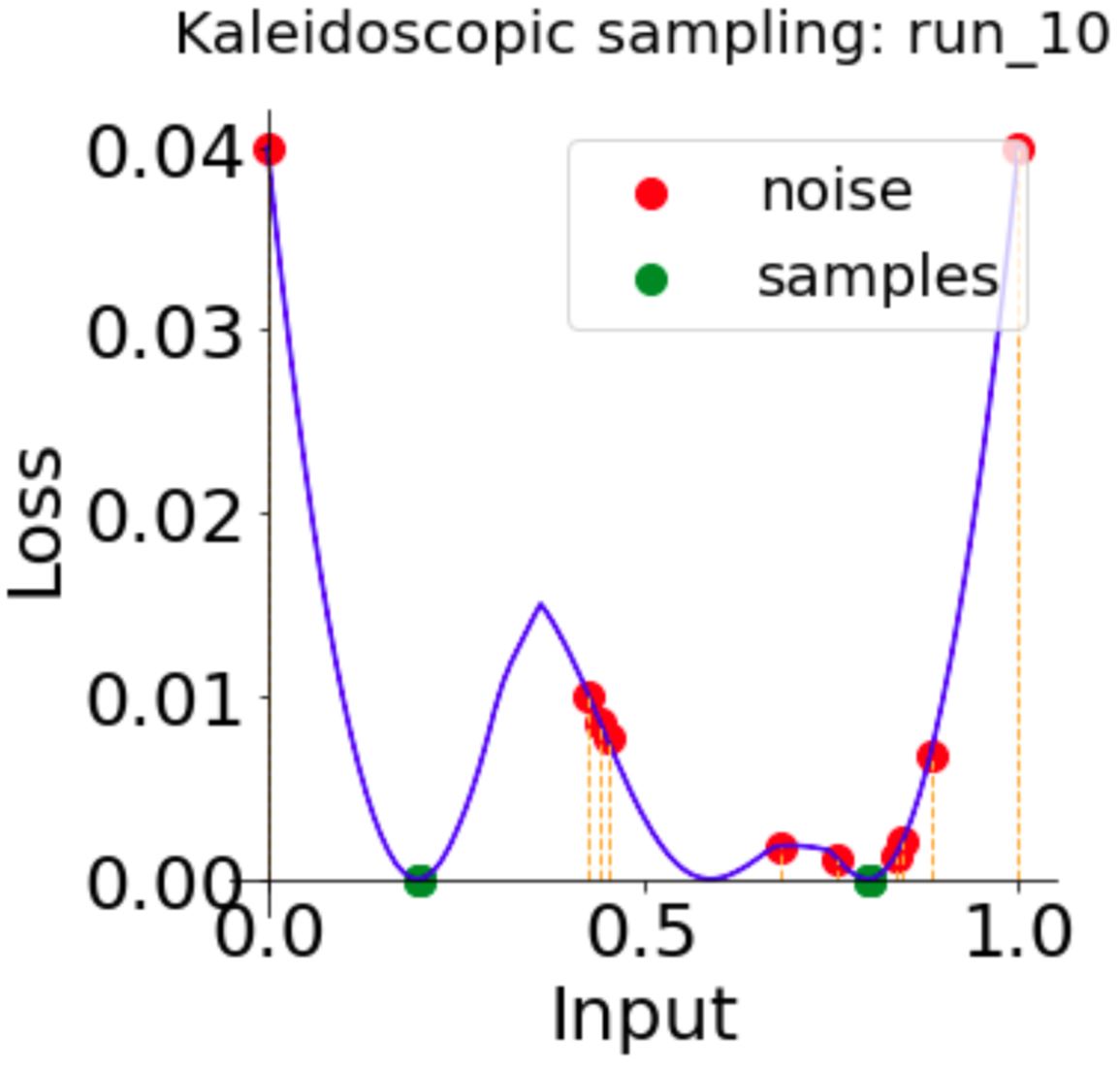}}
\caption{\small \textbf{Multiple points in 1D space}. We run manifold learning on data points $X=\{0.2, 0.8\}$ by fitting a MLP with $H=5,L=2$ (top 2 rows) and $H=5,L=7$ (bottom 2 rows) and in each case, we run for epochs $\geq 1K$ to ensure that the training loss tends to $\rightarrow 0$. Intermediate layers had `ReLU' non-linearity and `Sigmoid' at the final layer. By design, the manifold learning is supposed to match the input and output. But, on the contrary, we can observe the over-generalization phenomenon in (b,h), where the MLP learns many-to-one mapping, as evident by the flat regions around the points $\{0.2, 0.8\}$. Scaled version of loss shown in (a,g) on the zoomed in part as the increasing and spreading the training points across the space makes the loss manifold flatter. The rows (c-f) and (i-l) shows intermediate instances of the sampling runs. The burn-in period is roughly around $B=5$, which indicates that the sampling converges fast. The loss manifold is relatively flat between 0.2 to 0.8 and that causes the samples to converge slower while running the sampling procedure. As compared to (b), in (h) we can find an increase in the flat region around input values $0.2$ and $0.8$, which in turn makes the curve connecting the flat regions more steeper and thus our sampling procedure works faster. As this ensures that the outputs (y-axis) are mapped closer to the training inputs points with increasing number of iterations of the \kals sampling procedure. (best viewed in color)}
\label{fig:multiple-pts-1D}
\end{figure}

\textit{1. Analysing a single point in 1D space}: Fig.~\ref{fig:single-point-1D} (a) considers manifold learning on a single one-dimensional data point, i.e. we learn the mapping such that the input matches the output. The loss should be low only on the points that the MLP has been trained on and should be high for all the other points in the space. Based on the loss formulation in Eq.~\ref{eq:manifold-learning}, the max loss value can be 1. The plot of output vs the input in Fig.~\ref{fig:single-point-1D}(b) verifies that the over-generalization phenomenon premise that the output values of a learned neural network are limited to the range of output values observed during training. In an alternative scenario, which does not follow the over-generalization phenomenon, one might expect the loss at points other than $x=0.5$ to be closer to 1. 

\textit{\textbf{Intriguing question}: How to make the MLP simply learn an identity function? It is theoretically possible as the number of hidden units is greater than the input dimension and we just have a single training data point.} [We understand the the initialization plays a key role for the behaviour of NN over the untrained input values. This does open up a question on how to design the training.]

\begin{figure}
\centering 
($H=10, L=10$)
\subfigure[]{\includegraphics[width=0.5\textwidth]{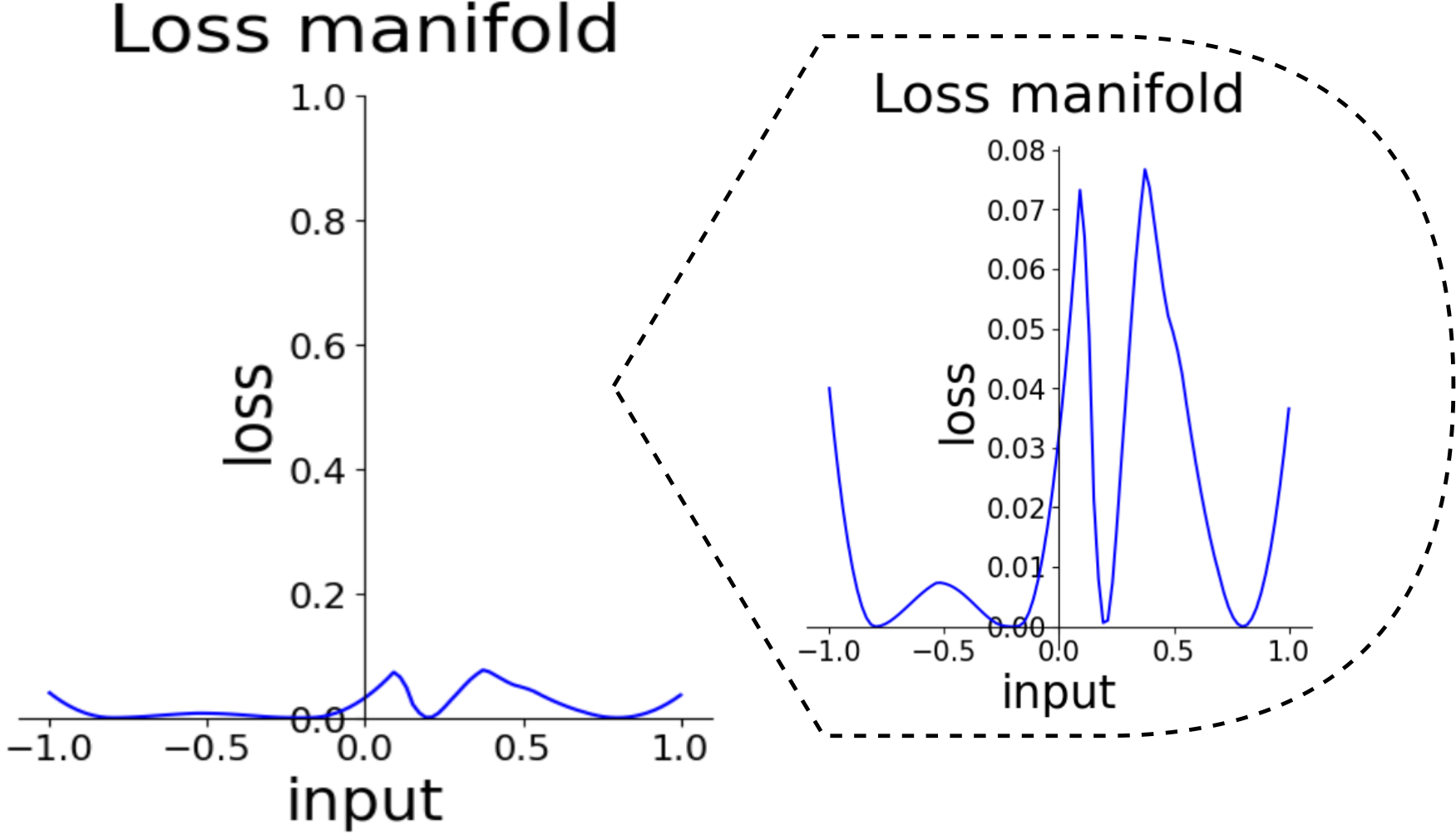}~}\quad
\subfigure[]{\includegraphics[width=0.25\textwidth]{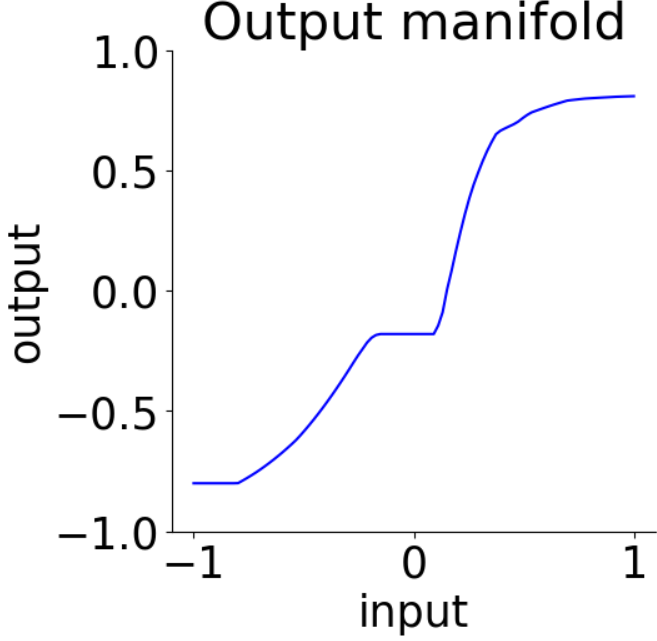}~}
\subfigure[]{\includegraphics[width=0.24\textwidth]{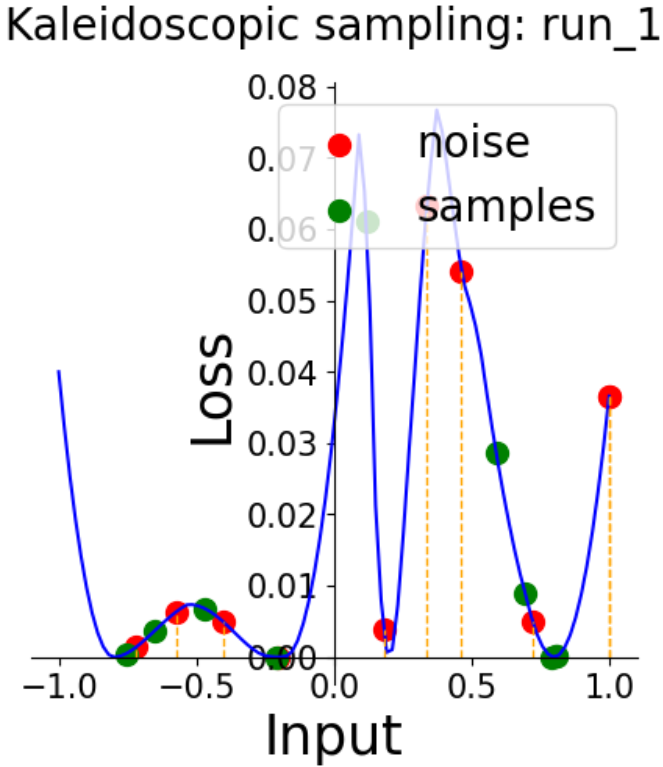}~}
\subfigure[]{\includegraphics[width=0.24\textwidth]{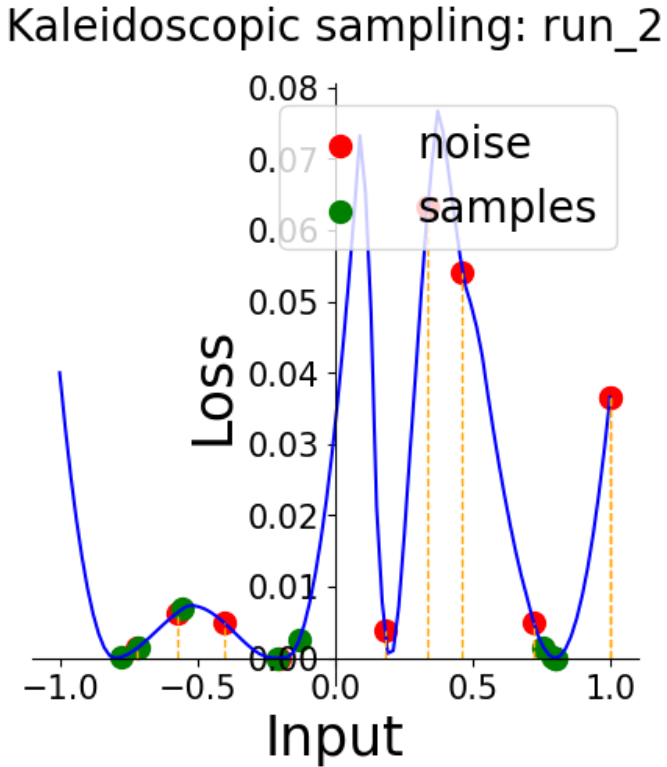}~}
\subfigure[]{\includegraphics[width=0.24\textwidth]{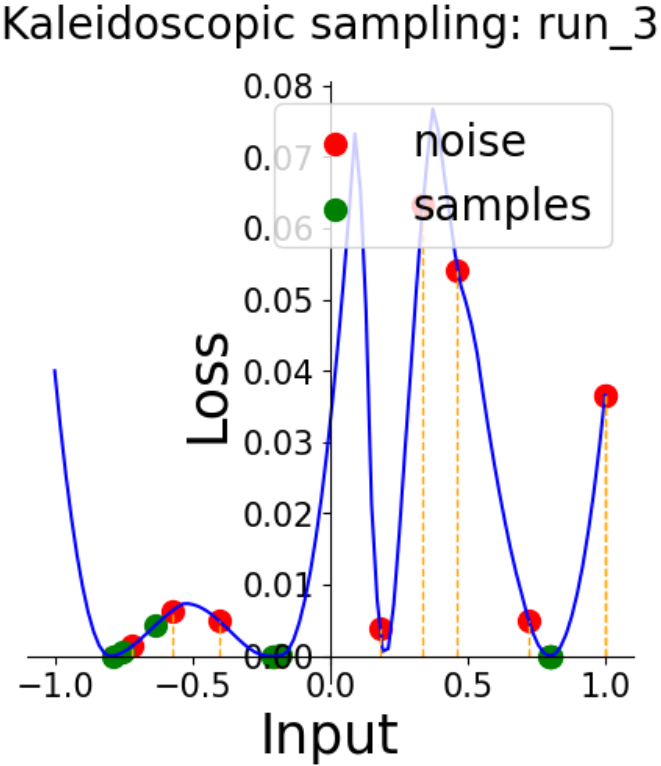}~}
\subfigure[]{\includegraphics[width=0.24\textwidth]{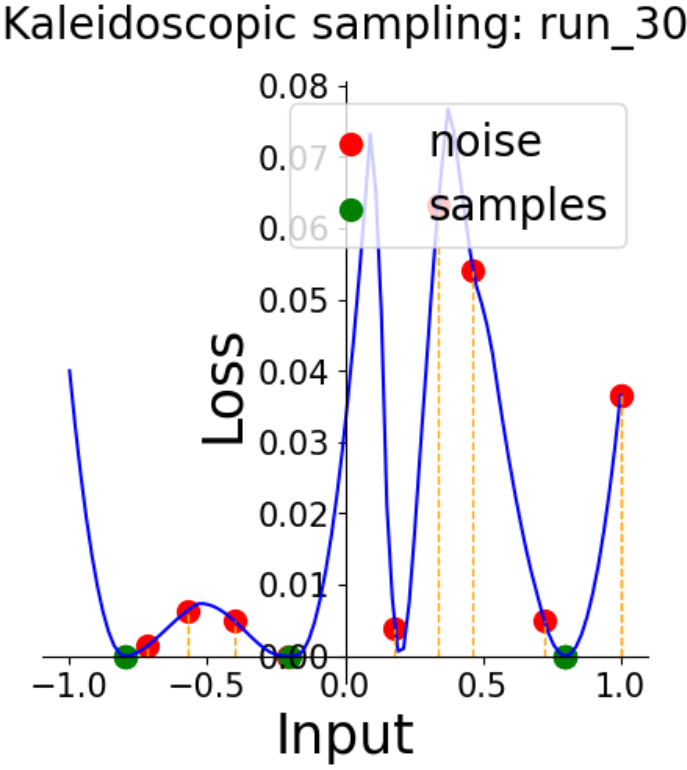}}
\caption{\small \textbf{Multiple points in 1D space with NN bounded by `Tanh'}. We run manifold learning on data points $X=\{-0.8, -0.2, 0.2, 0.8\}$ by fitting a MLP with $H=10,L=10$ and the final layer non-linearity as `Tanh'. This basically expands the range of the input and output between $[-1,1]$. The rows (c-f) shows intermediate instances of the sampling runs. We can observe the over-generalization phenomenon in (b), where the MLP learns many-to-one mapping, as evident by the flat regions around the points $X$. (best viewed in color)}
\label{fig:multiple-pts-1D-tanh}
\end{figure}

\textit{2. Analysing multiple points in 1D space}: Things start to get interesting as we increase the number of points. We do manifold learning with a MLP on points $X=\{0.2, 0.8\}$ and observe the loss and output manifolds, refer Fig.~\ref{fig:multiple-pts-1D}. In this experiment, we also highlight the effect of \textit{\textbf{over-parameterizing neural networks}}, refer to the contrast between top and bottom settings in Fig.~\ref{fig:multiple-pts-1D}, where we increased the number of parameters from $H=5, L=2$ to $H=5, L=7$. We found that, in general, the output manifold over the range of the input values becomes more restricted with increasing the complexity of the MLP and thus require less number of sampling runs for convergence. Fig.~\ref{fig:multiple-pts-1D-tanh} shows similar exercise with more number of points using `Tanh' as the choice of non-linearity. 

\begin{figure}
\centering 
\subfigure[x=(0.5, 0.5)]{\includegraphics[width=0.45\textwidth]{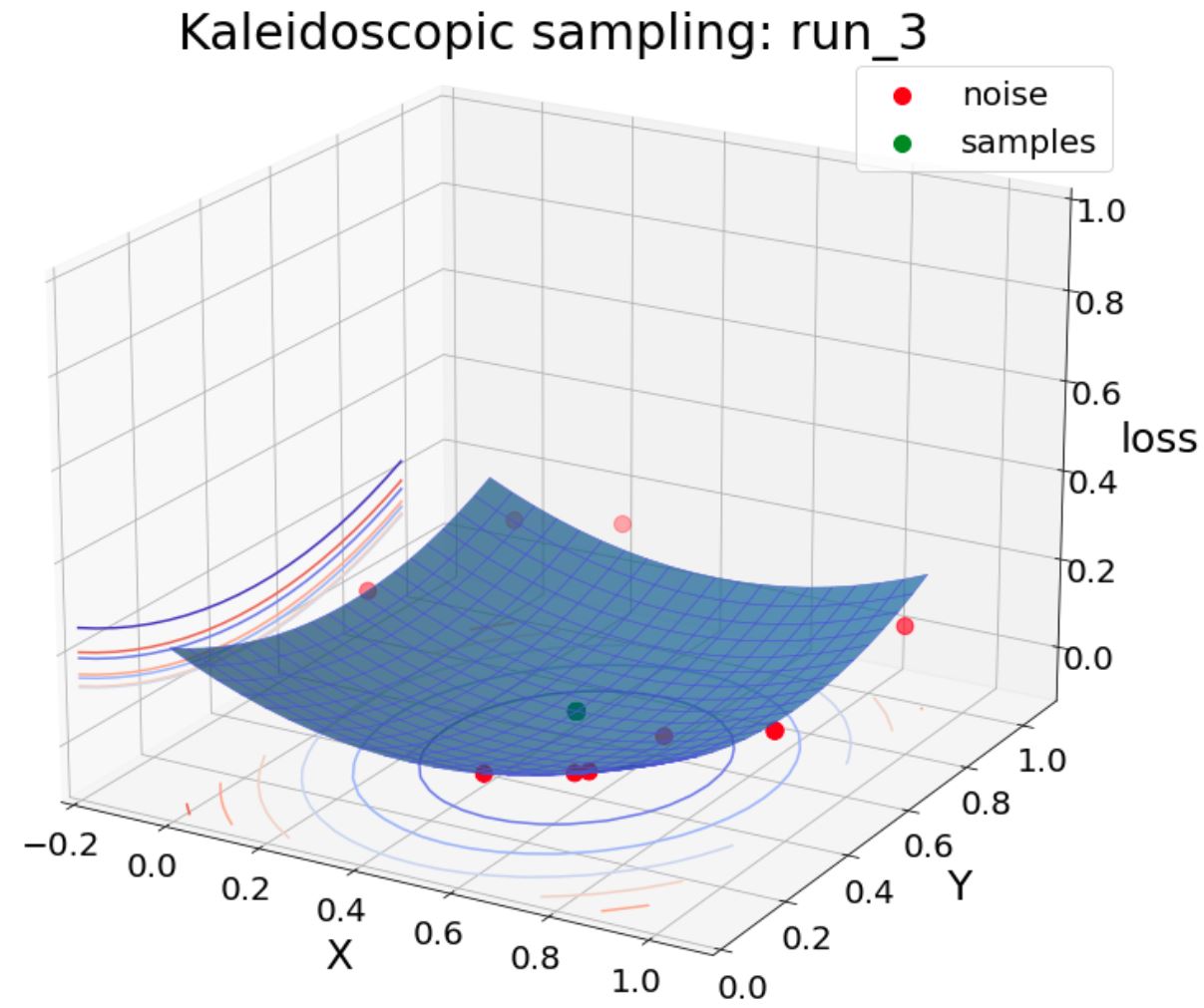}~~\quad} 
\subfigure[\text{\small x=[(0.2, 0.2), (0.2, 0.8), (0.8, 0.2), (0.8, 0.8)]}]{\includegraphics[width=0.45\textwidth]{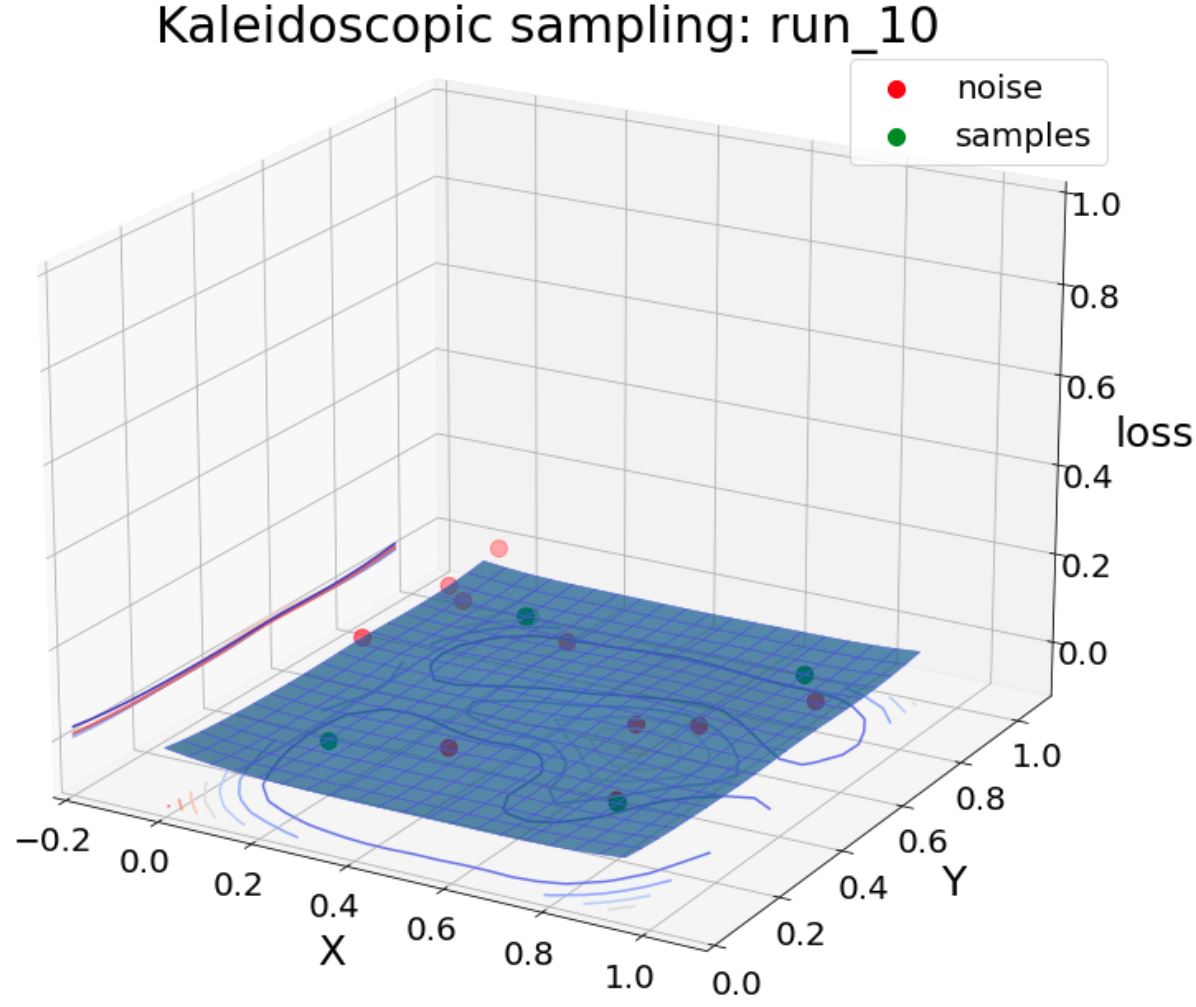}}
\caption{\small \textbf{Loss manifold and sampling in 2D space}: (a) We do manifold learning in 2-dimensions with a MLP $L=2, H=50$ at point $x=(0.5,0.5)$ and (b) a MLP $L=7, H=50$ at the points $x=[(0.2,0.2), (0.2,0.8), (0.8,0.2), (0.8,0.8)]$. Each plot initializes a random noise (in {\color{bittersweet} red}) sampled from a Normal distribution $\mathcal{N}(0.5, 0.5I)$ and ran \kals sampling whose samples are shown in {\color{aoenglish}green}. The loss function hyperplane is quite flat but still a large number of samples are obtained near the input distribution which indicates that our sampling procedure is working which in turn suggests the existence of the `over-generalization' phenomenon. We found that the number of sampling runs for reaching the burn-in period is inversely proportional to the complexity of the neural network. Note that the scatter points can look slightly off the manifold due perspective angle adjustments in 3D rendering. (best viewed in color)}
\label{fig:analysis-pts-2D}
\end{figure}

\textit{3. Analysing points in 2D space}: Going to higher dimensions and making inference from the results is tricky in general, as we lose our ability to visualize the data. Fig.~\ref{fig:analysis-pts-2D}, shows that the over-generalization phenomena is prominent even as we go towards higher dimensions. 

We can see that \textit{Neural networks learn a concentrated or many-to-one mapping}. Basically, the output for the unobserved points are concentrated to the
output ranges that were generated for the training data. we utilize this phenomenon to design an algorithm that can recover `meaningful' samples.

\subsection{Extrapolating neural network's behavior in higher-dimensions}\label{sec:nn-high-dim}

\begin{figure}
\floatbox[{\capbeside\thisfloatsetup{capbesideposition={left, center},capbesidewidth=8.0cm}}]{figure}[0.999\FBwidth]
{
\subfigure{\includegraphics[height=55mm]{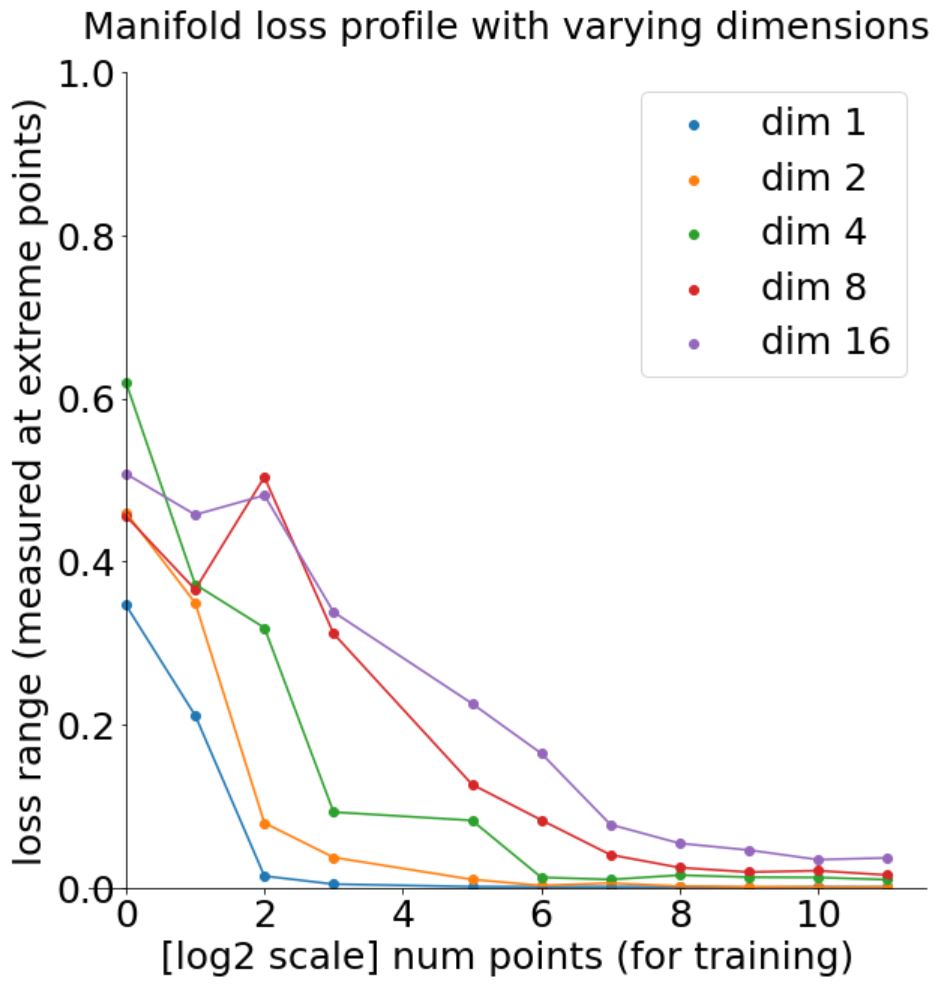}}}
{\caption{\small \textbf{Loss manifolds in higher dimensions}. As observed in Fig.~\ref{fig:single-point-1D}, the loss manifold is usually flatter in the middle than at the extreme points of the space. For a given dimension $D$, there are $2^D$ extreme points. Say, if the space is $\mathbb{R}^4$, then the extreme points of the space are $[(0,0),(0,1),(1,0),(1,1)]$. We first initialize some points uniformly at random on the $\mathbb{R}^D$ space and run manifold learning till convergence (min loss$\rightarrow0$). On the x-axis, we plot the number of training points which are chosen as powers of 2, from $2^0\rightarrow 2^{11}$. The model chosen was a MLP with $L=2, H=50$. To get the loss range, we evaluate the loss at the extreme points of the space and choose the max loss value. For $R^{16}$ space, we evaluated the loss at $2^{11}=65536$ extreme points. A \textit{\textbf{key takeaway}} here is that if the number of training points are large, then no matter the dimension of the input space, the manifold learned by the neural networks will become very flat with respect to the loss values over the entire input range.}\label{fig:manifold-loss-vs-num-pts}}
\end{figure}

\textit{Escaping the `curse' of higher dimensions}. Fig.~\ref{fig:manifold-loss-vs-num-pts} measures max loss values for varying number of datapoints in higher dimensional spaces. As the loss manifold becomes more flat with increasing number of points over different input data dimensions, we extrapolate and assume that their will be more local minima in the regions outside of the input distribution on which the model was trained. Or the average loss value of the entire manifold (i.e. loss over the whole input range and not just the training data) decreases with the increase in number of points and their spread. Thus, it will become increasingly difficult to get high-fidelity samples, as the curve flattens causing the probability of noisy samples to increase. Although, a point in our favor is that the `\textit{\textbf{flatness of the loss manifold is proportional to the spread of the data points over the space regardless of the input dimension size}}'. This helps narrow down our strategy for designing a robust sampling procedure.

\textit{Validating the `dimple manifold' concept}. We digress a bit and address an important observation made in this work~\cite{shamir2021dimpled} about NNs learning dimple manifolds. Based on the loss function vs number of points curve over different dimensions of input, refer Fig.~\ref{fig:manifold-loss-vs-num-pts}, we can experimentally say that the dimple manifold proposal that the NNs hyperplane creates a `dimple' (or a dip, or a small trench) around the training data points. Hence, with slight change in the NN weights, one can change the NN output for a given input data point. As the loss function curve is relatively flat, there is not much loss difference between the observed (training) and the unobserved (testing) data points. 
This also validates the general observation that one can adversarily attack/trick the neural network by slightly perturbing the input. From the Fig.~\ref{fig:single-point-1D},\ref{fig:multiple-pts-1D},\ref{fig:multiple-pts-1D-tanh}\&\ref{fig:analysis-pts-2D}, we can see that a \textit{smooth manifold dip} is formed on the loss manifold at the training points. Now, if there are many such training points, then the plots in Fig.~\ref{fig:manifold-loss-vs-num-pts} suggests that the manifold will have a very low loss values throughout the space. Thus, when adversarily attacking the model, one can make slight modifications to the input to increase its loss value and thereby the point is pulled out from the manifold dip and gets misclassified.




\section{Generative \textbf{\kals} Networks}

\textit{Objective}: Given a set $X\in\mathbb{R}^{M\times D}$ with M samples of D dimensions each, we want to generate similar samples coming from the underlying joint distribution that represents the input data. 

\textit{Intuition}. If the manifold learning model $f_\mathcal{N}$ is properly trained by ensuring that the loss function is sufficiently minimized, we will have high probability of input points $X$ to have lower loss values. So, if we figure out a way to generate a point with lower loss, then it can be considered a valid sample. A key issue is to learn a joint probability distribution over the input data space as we do not know the probabilities, or difficult to set the probabilities, of the points that are not seen during training. 

\textit{Approach}. The `over-generalization' phenomena suggests that the output is restricted to the output range as seen during training of a bounded neural network. We utilize this phenomenon to generate samples from the training data distribution. Specifically, while learning we make the output match to the input of the neural network (e.g. a reconstruction setting). Now, if we give any random input to the neural network, it will map to the output range, which in turn is the same as the input range. Now, if we again apply the same neural network to the observed output value, it will map it further into the input distribution region. We keep continuing this process and after a certain burn-in period, we find that the samples converge. Fig.~\ref{fig:manifold-learning-sampling}[right] demonstrates this procedure. Additionally, from our observation about the loss manifolds in higher dimensions, the behavior of \kals sampling procedure on the manifold learned by MLPs, refer Eq.~\ref{eq:manifold-learning}, will not be affected by the dimension of the input. 

\textbf{\kals sampling}: The learned neural network model weights $f_\mathcal{N}$ are frozen, refer Eq.~\ref{eq:manifold-learning}. An random input tensor is initialized either from a Uniform distribution or a Normal distribution, e.g. $z\sim\mathcal{U}(0, 1)$. The model $f_\mathcal{N}$ is then repeatedly applied till burn-in period achieved in $B$ iterations. After that any further applications of the model will lead to converged samples. 
\begin{align}\label{eq:manifold-sampling}
     f^B_\mathcal{N}(z) = \{f_\mathcal{N}\circ \cdots (\text{B times})\cdots \circ f_\mathcal{N}\}(z) 
\end{align}
In practice, to get an enhanced `Kaleidoscopic' behaviour, we add a bit of noise to each iteration. This facilitates jumping between the steps (or different data points). 
\begin{align}\label{eq:manifold-sampling-eps}
      x^1 &= f_\mathcal{N}(x^0) + \epsilon^0\\\nonumber
      &\cdots (\text{B times})\\\nonumber
      x^B &= f_\mathcal{N}(x^{B-1}) + \epsilon^B\nonumber
\end{align}
where $\epsilon$ is small noise sampled from either a normal or uniform distribution. 

\textit{Choice of neural network}. For Multi-layer Perceptrons, we found that increasing the complexity in terms of depth (number of hidden layers) works better than increasing the complexity in terms of the width (number of hidden units) of the MLP. We observed `{\textit{strong positive correlation between the over-generalization phenomenon and the depth of the MLP}}' as we get significantly improved sampling results. As we want the output manifold to follow step-wise behaviour, our observation of using `deep ReLU' networks matches the Kolmogorov-Arnold representation modifications discussed in ~\cite{schmidt2021kolmogorov} (Note, we are still investigating this connection). 

\begin{figure}
\centering 
\subfigure{\includegraphics[width=0.3\textwidth]{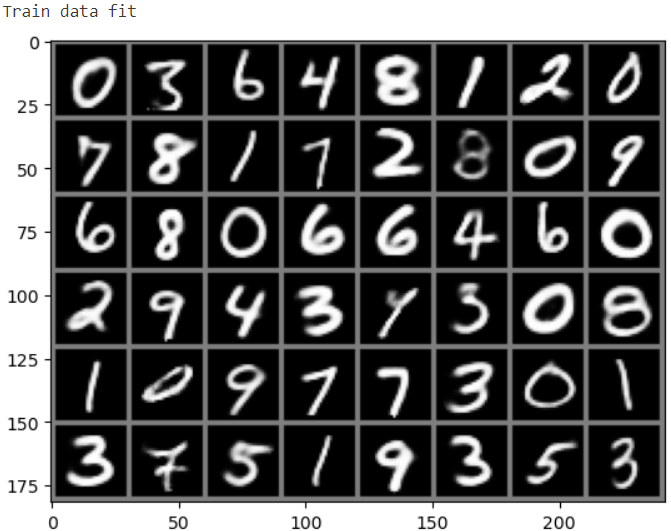}~} 
\subfigure{\includegraphics[width=0.3\textwidth]{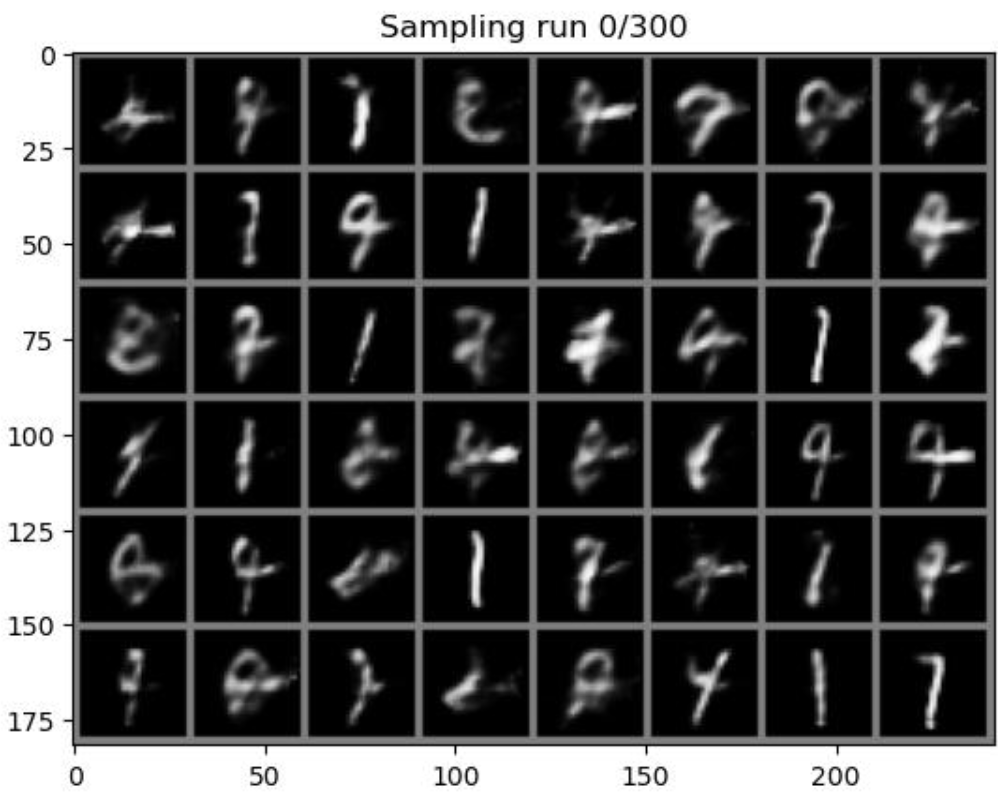}} 
\subfigure{\includegraphics[width=0.3\textwidth]{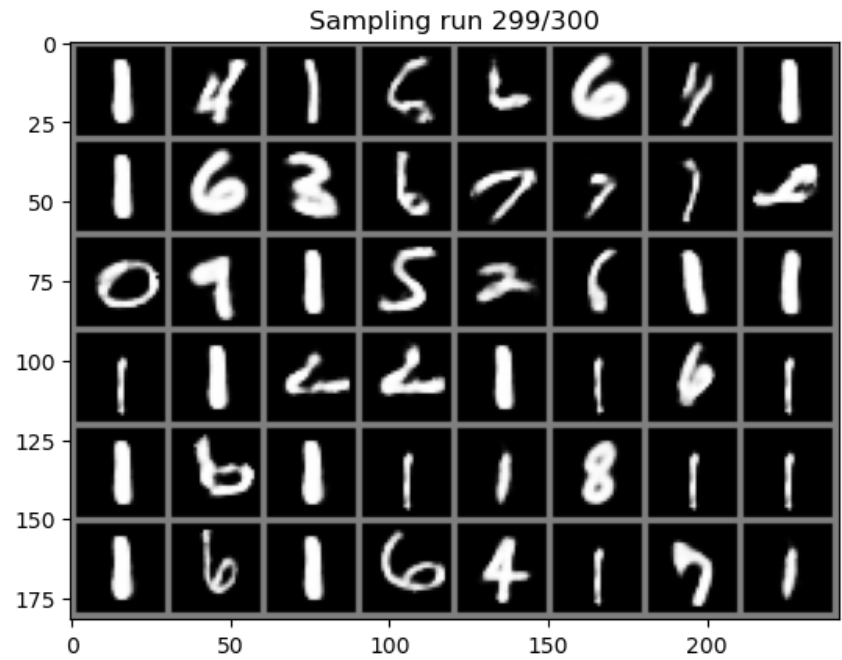}}
\subfigure{\includegraphics[width=0.18\textwidth]{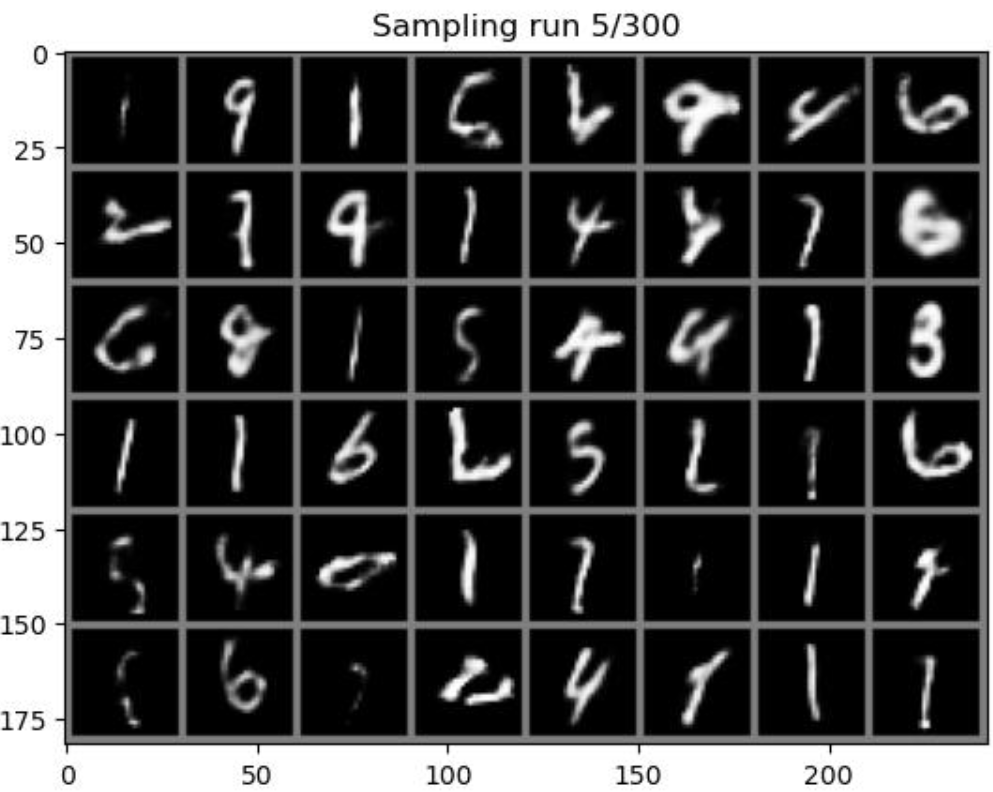}}
\raisebox{-4.5\height}{\subfigure{\includegraphics[height=2.5mm, width=1.5mm]{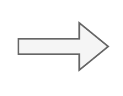}}}
\subfigure{\includegraphics[width=0.18\textwidth]{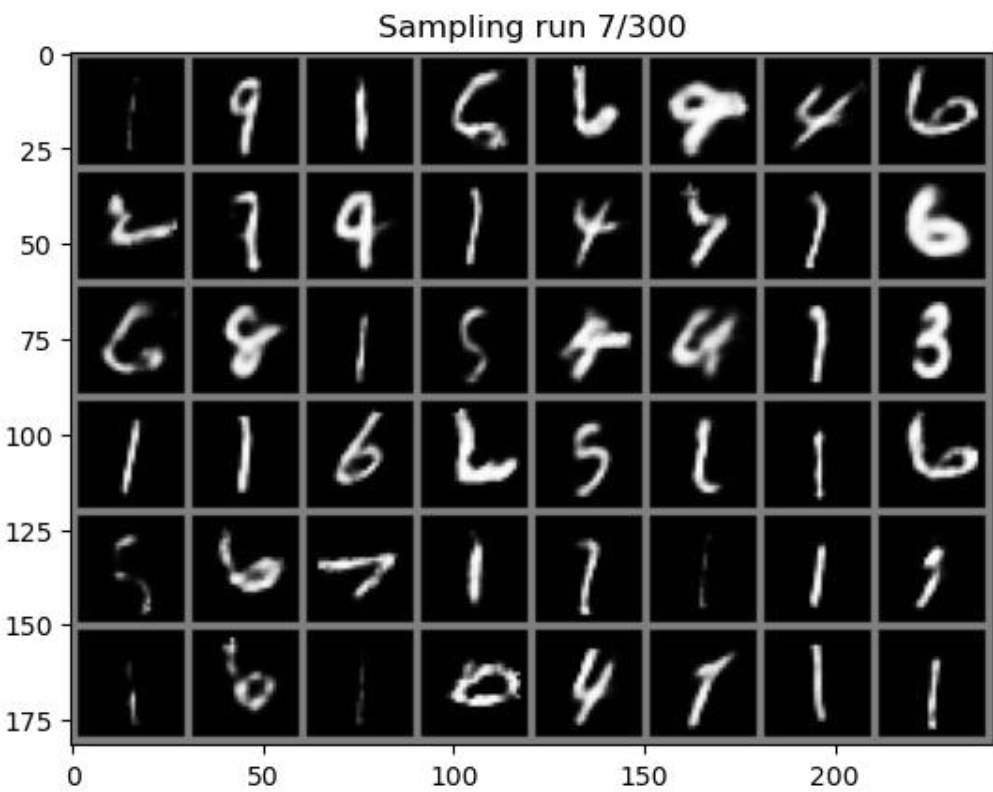}}
\raisebox{-4.5\height}{\subfigure{\includegraphics[height=2.5mm, width=1.5mm]{kals_figures/right_arrow.pdf}}}
\subfigure{\includegraphics[width=0.18\textwidth]{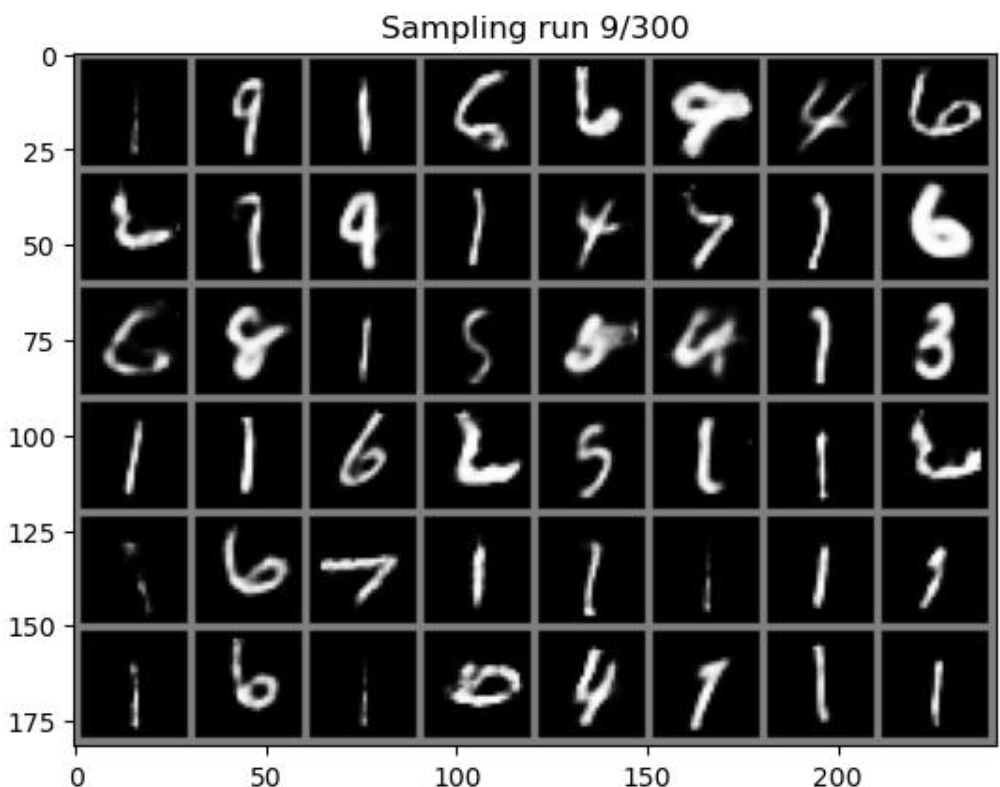}}
\raisebox{-4.5\height}{\subfigure{\includegraphics[height=2.5mm, width=1.5mm]{kals_figures/right_arrow.pdf}}}
\subfigure{\includegraphics[width=0.18\textwidth]{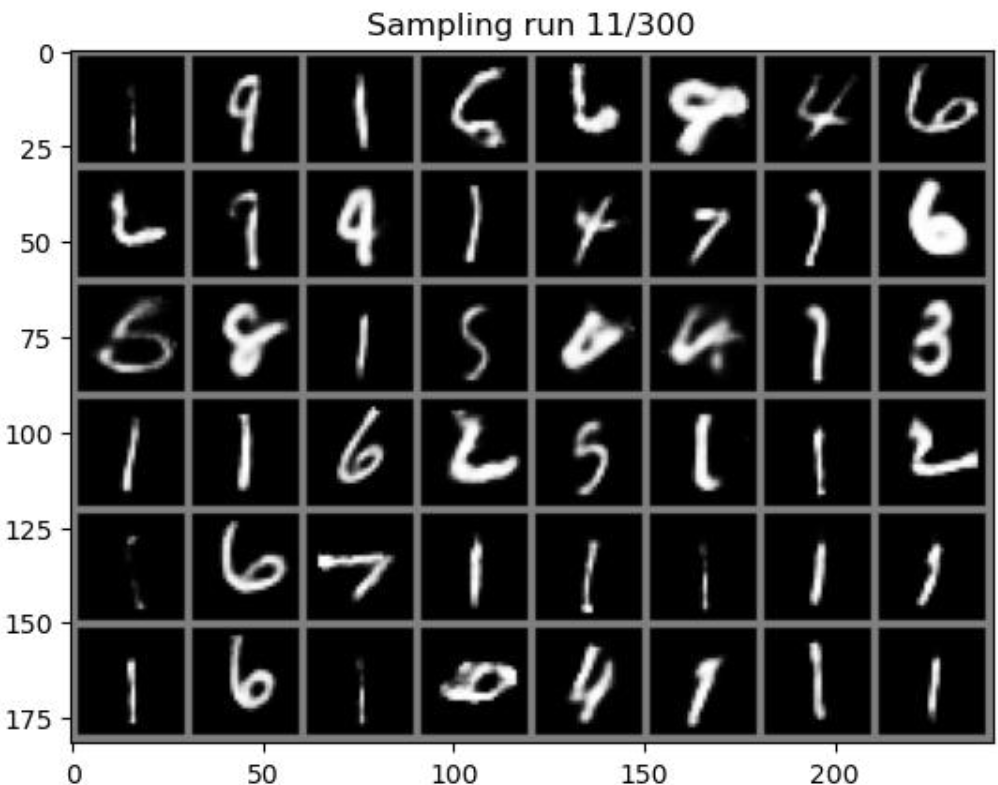}}
\raisebox{-4.5\height}{\subfigure{\includegraphics[height=2.5mm, width=1.5mm]{kals_figures/right_arrow.pdf}}}
\subfigure{\includegraphics[width=0.18\textwidth]{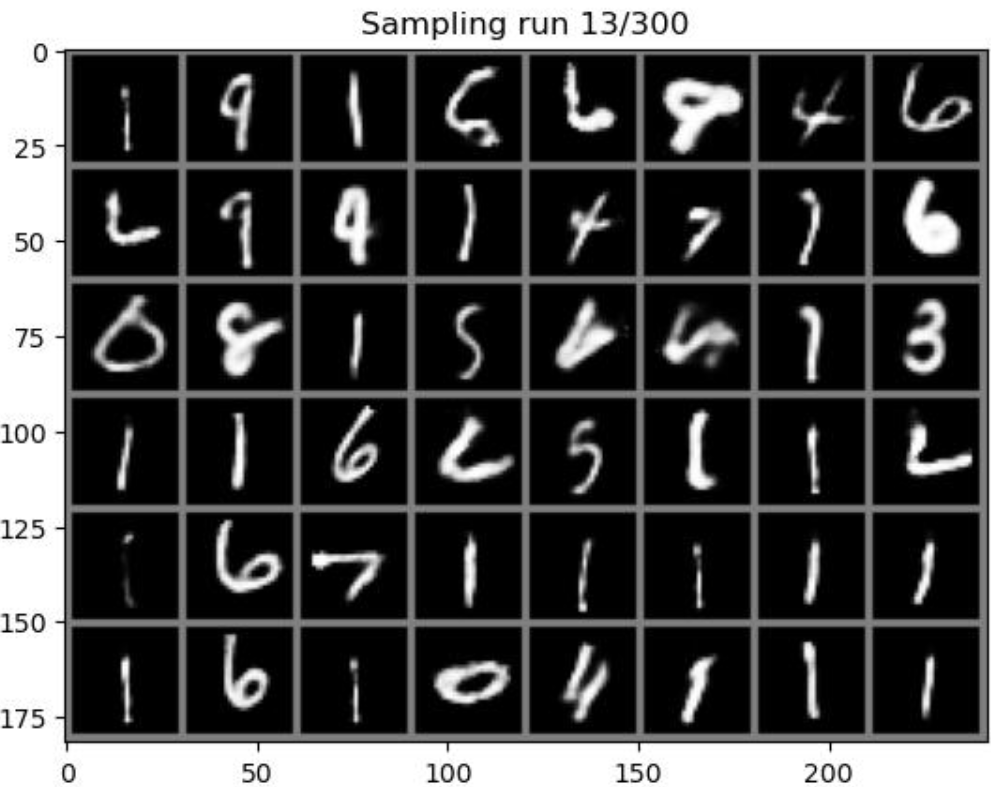}}
\subfigure{\includegraphics[height=10mm]{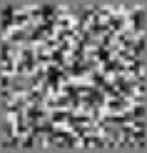}}
\subfigure{\includegraphics[height=10mm, width=7mm]{kals_figures/right_arrow.pdf}}
\subfigure{\includegraphics[height=10mm]{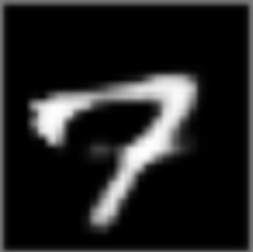}}
\subfigure{\includegraphics[height=10mm, width=7mm]{kals_figures/right_arrow.pdf}}
\subfigure{\includegraphics[height=10mm]{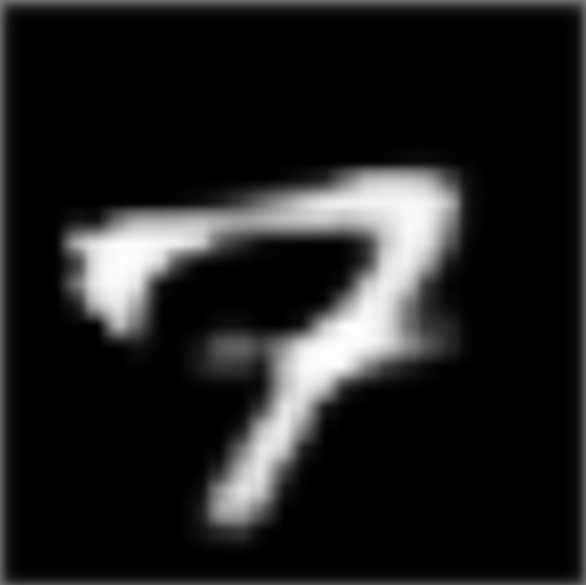}} 
\subfigure{\includegraphics[height=10mm, width=7mm]{kals_figures/right_arrow.pdf}}
\subfigure{\includegraphics[height=10mm]{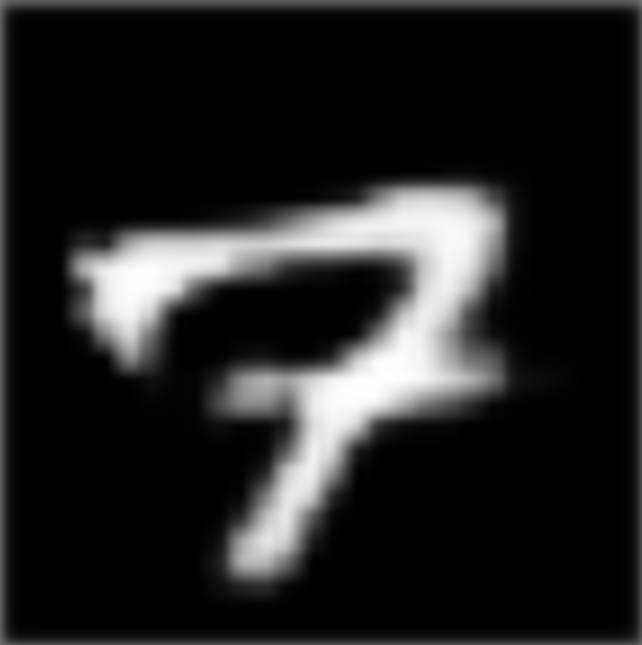}}
\subfigure{\includegraphics[height=10mm, width=7mm]{kals_figures/right_arrow.pdf}}
\subfigure{\includegraphics[height=10mm]{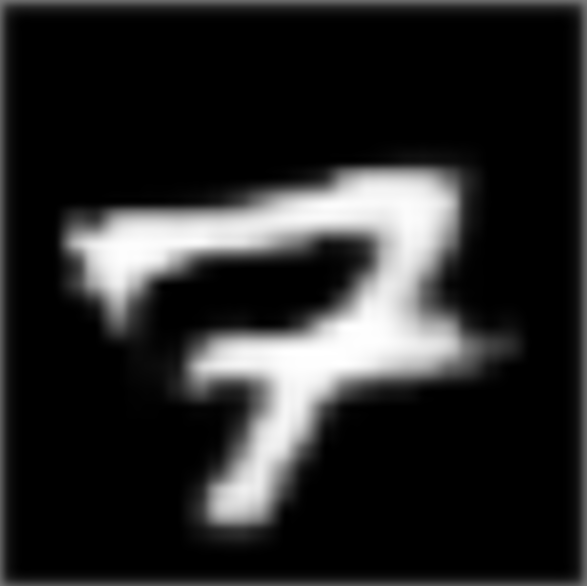}} 
\subfigure{\includegraphics[height=10mm, width=7mm]{kals_figures/right_arrow.pdf}}
\subfigure{\includegraphics[height=10mm]{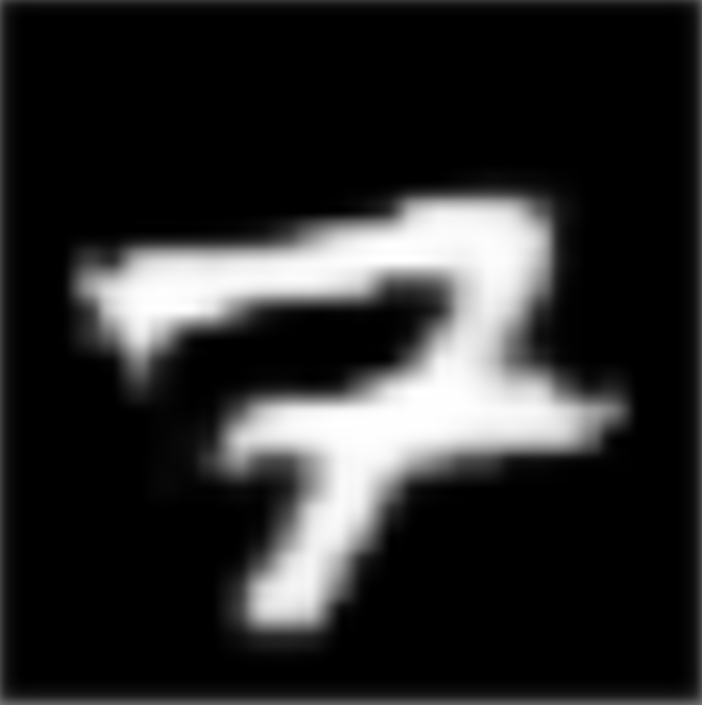}}
\caption{\small \textbf{Generative Kaleidoscopic Network for MNIST}: We do manifold learning on a MLP with $L=10, H=500$ with intermediate `ReLU' and final layer with `Tanh' non-linearity. We create a `MNIST Kaleidoscope' by setting $\epsilon=0.01$ in the \kals sampling procedure as described in Eq.~\ref{eq:manifold-sampling-eps}. [TOP row] The leftmost images show the digits recovered after the manifold learning. The center images show the $1^{st}$ run after applying learned MLP on the input noise. The input noise vector was sampled randomly from a Uniform distribution $~\mathcal{U}(-1,1)$. (Note, we get similar results with Normal distribution too). The rightmost images show the state at the sampling run of 300. We note that , at times, the procedure can converge at a digit and then it can remain stable throughout the future iterations as it has found a stable minima or step as defined in our analysis. For this reason, we add a slight noise at every sampling iteration to simulate kaleidoscopic effect. Here, we show the output of randomly chosen subset of images and the seed is constant, so that there is one-to-one correspondence across sampling runs. [MIDDLE row] A sequence of intermediate sampling runs $5\rightarrow13$, which is still in the burn-in period. One can take a digit and observe their evolution over the iterations. For instance, top row \& third last column evolves into digit $8$ and then eventually transitions to digit $6$ at the $300^{th}$ iteration. [BOTTOM row] Shows another sequence of evolution of sevens from noise.}
\label{fig:kals-MNIST-MLP}
\end{figure}

\textit{Observations}: We revisit the Fig.~\ref{fig:single-point-1D},\ref{fig:multiple-pts-1D},\ref{fig:multiple-pts-1D-tanh}\&\ref{fig:analysis-pts-2D} which analyze points in 1D and 2D spaces. We observe from the loss manifold profiles that one can get also get multiple minima in the regions where the training data is not present. This means that the probability of getting samples from that region becomes high, which is not desired. When we see the corresponding output manifold profiles and sampling results, we can see that if we repeatedly apply the model $f_\mathcal{N}$, one can get samples from the training data distribution but will require more number of iterations (or longer burn-in periods). The \kals sampling procedure will improve if we can make the output manifold look like a \textbf{\textit{step-function}} over the points observed in training distributions which happens when we increase the depth of the MLP. Fig.~\ref{fig:kals-MNIST-MLP} shows the \kals sampling results for MNIST images. In our mind, we visualize that the output vs input profile of the MLP as multiple steps (many-to-one mappings), with each step containing the distribution of a digit. Thus, we can start with any input random noise and end up on one of these digit steps after the burn-in period of the \kals sampling. 

\textit{Limitations}: We were able replicate, albeit with limited success, the \kals samples with other deep learning architectures for generating images. We tried architectures like the variants of Convolutional Neural Networks, Transformers and U-Nets on different image datasets like the MNIST \& CIFAR-10. We observed that our sampling procedure does converge but the samples do not represent the images from the training data used in the manifold learning process. We believe that the over-generalization phenomenon (or the many-to-one mapping) still holds for these architectures but it might be creating additional unwanted `steps' in the output vs input profile. This may occur due to the inherent bias in the deep learning architecture chosen. Still, we believe that more experimentation is needed to understand the reasons behind this behaviour. 

\begin{figure}
\centering 
\raisebox{-0.6\height}{\subfigure{\includegraphics[width=0.17\textwidth]{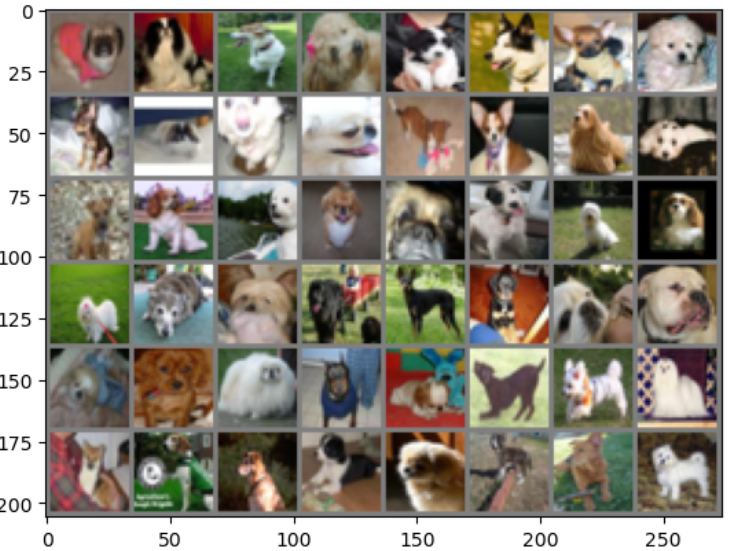}~}}
\subfigure{\includegraphics[width=0.005\textwidth, height=20mm]{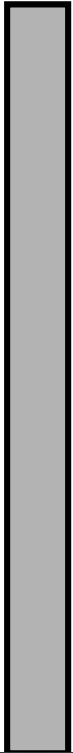}~}
\subfigure{\includegraphics[width=0.18\textwidth]{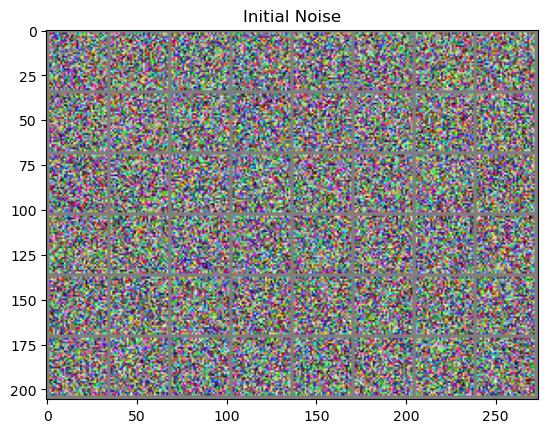}}
\raisebox{-4.5\height}{\subfigure{\includegraphics[height=3mm, width=2mm]{kals_figures/right_arrow.pdf}}}
\subfigure{\includegraphics[width=0.18\textwidth]{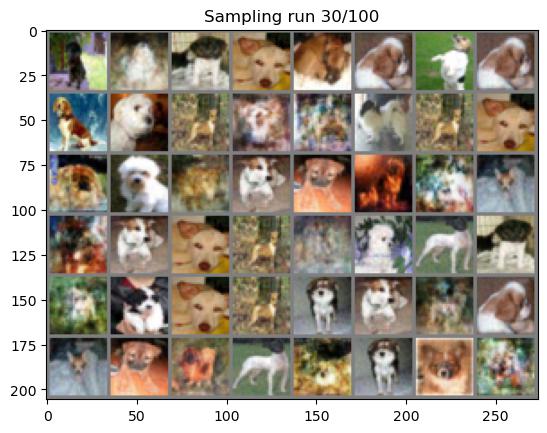}}
\raisebox{-4.5\height}{\subfigure{\includegraphics[height=3mm, width=2mm]{kals_figures/right_arrow.pdf}}}
\subfigure{\includegraphics[width=0.18\textwidth]{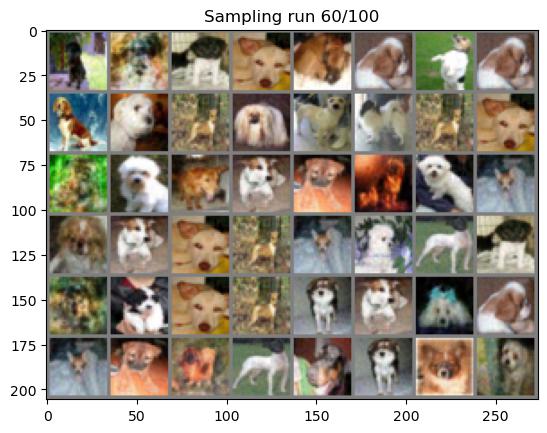}}
\raisebox{-4.5\height}{\subfigure{\includegraphics[height=3mm, width=2mm]{kals_figures/right_arrow.pdf}}}
\subfigure{\includegraphics[width=0.18\textwidth]{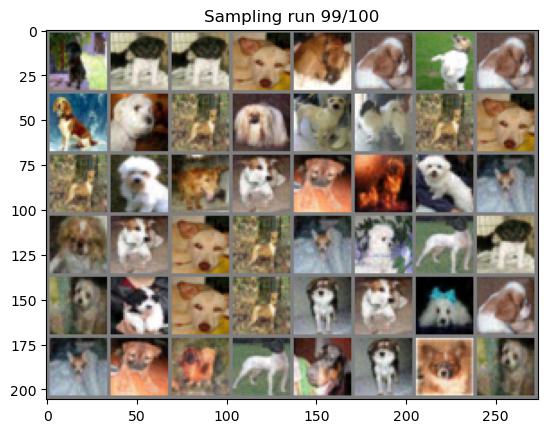}}
\raisebox{-0.8\height}{\subfigure{\includegraphics[width=0.17\textwidth]{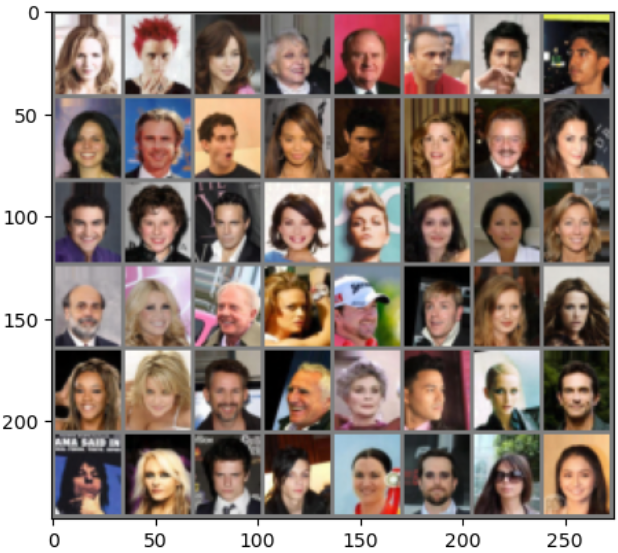}~}}
\subfigure{\includegraphics[width=0.005\textwidth, height=24mm]{kals_figures/vertical_divider}~}
\subfigure{\includegraphics[width=0.18\textwidth]{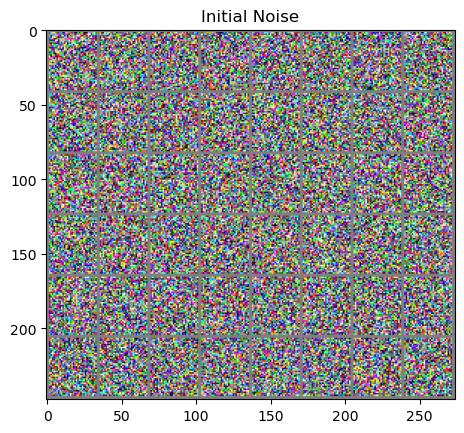}}
\raisebox{-5.5\height}{\subfigure{\includegraphics[height=3mm, width=2mm]{kals_figures/right_arrow.pdf}}}
\subfigure{\includegraphics[width=0.18\textwidth]{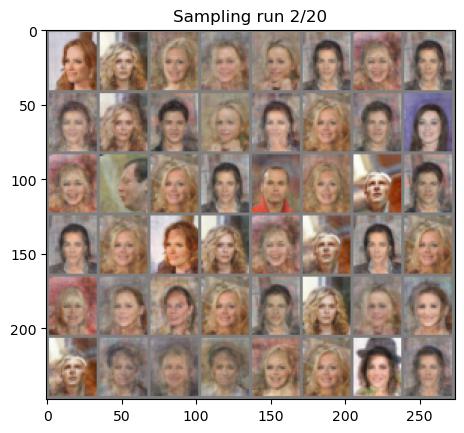}}
\raisebox{-5.5\height}{\subfigure{\includegraphics[height=3mm, width=2mm]{kals_figures/right_arrow.pdf}}}
\subfigure{\includegraphics[width=0.18\textwidth]{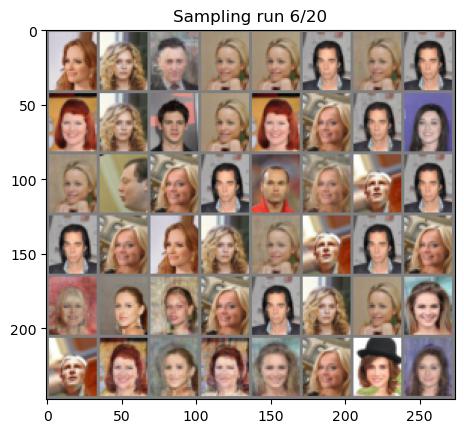}}
\raisebox{-5.5\height}{\subfigure{\includegraphics[height=3mm, width=2mm]{kals_figures/right_arrow.pdf}}}
\subfigure{\includegraphics[width=0.18\textwidth]{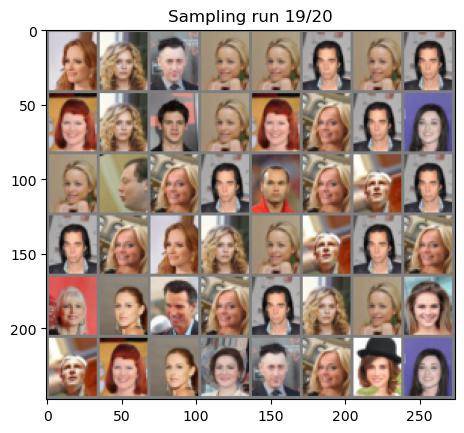}}
\caption{\small \textbf{Generative Kaleidoscopic Networks: CIFAR \& CELEB-A}. We create dataset kaleidoscopes on a subset of the images. We do manifold learning with MLP $L=10, H=2000$. [TOP] We randomly chose 1K dog images from CIFAR-10 dataset. The leftmost image shows some images from the training data, while the remaining images are the samples obtained during the intermediate sampling transitions starting with initial random noise. [BOTTOM] We randomly chose 1K celebrity images from the Celeb-A dataset.  In this case, we observe more `jumpy' transitions between sampling and the convergence in term of the sampling iterations also occurs quicker. Our initial reasoning for this is that the image manifold is not smooth and hence there are more steps in the output manifold causing the sampling procedure to jump abruptly or get stuck at a sample. (Animations included in the demo link)}
\label{fig:kals-cifar-celeba-mlp}
\end{figure}

\textit{Are we just memorizing? \textbf{Depends on the Dataset Manifold.}} The deeper the MLP, the stronger is the many-to-one mapping, which in turn makes the output vs input manifold look more like a step function. 
For MNIST images, refer Fig.~\ref{fig:kals-MNIST-MLP}, we get decent loss convergence and we visually can observe the `smooth' transitions among the samples recovered. We further measured distance from the closest training images and our initial conclusion suggests that it is not just purely memorizing. We further ran experiments on a subset of CIFAR and CELEB-A datasets, refer Fig.~\ref{fig:kals-cifar-celeba-mlp}. In some cases, we can observe some `jumpy' transitions between the \kals sampling iterations. Our initial assessment to explain this observations is that more steps in the neural network manifolds are formed if the dataset contains disconnected points and this indicates that the dataset is a collection from distinct distributions. In some sense, we can detect whether the dataset lies in a smooth manifold and whether there is `semantic' similarity among the datapoints. Another interesting aspect to explore is doing data compression using these networks. In order to help reader understand the sample transition reasoning discussed here, we added various `Dataset Kaleidoscopes' animations in the software link.

\section{Related Works}


\textit{Generative Modeling}: These surveys can be a good starting points for folks not familiar with the generative modeling paradigms like Autoencoder, Variational AE, Generative Adversarial Networks, Diffusion models and their variants~\cite{suzuki2022survey,cao2022survey}. In particular, it will be interesting comparison to re-visit the VAEs~\cite{kingma2013auto} continuous latent space generation in the light of \kals sampling over the AE architecture presented in this work. 

\textit{Properties of Neural Networks}: The paper on `Intriguing properties of neural networks'~\cite{szegedy2013intriguing} lists out some observations. Particularly interesting to us is this one ``We find that deep neural networks learn input-output mappings that are fairly discontinuous to a significant extent. We can cause the network to misclassify an image by applying a certain hardly perceptible perturbation, which is found
by maximizing the network’s prediction error.'' We modify this claim based on our findings for MLP architecture. \textbf{\textit{If we train deeper MLPs for images, it will become increasingly difficult to do adversarial attacks.}} As the step-length increases, refer Fig.~\ref{fig:multiple-pts-1D}(h), more perturbation will be needed to see changes in the output manifold. [Note, we were able to verify this behaviour for the MLP and hence, we cannot conclusively comment on the other deep learning architectures.] 

\textit{Gradient based sampling}: We consider the input to be a learnable tensor and then use a gradient descent based sampling procedure. Variants of this approach have been used previously as well in other contexts for doing adversarial attacks~\cite{xu2020adversarial} or learning and sampling probabilistic graphical models~\cite{shrivastava2023neural,shrivastava2022methods,shrivastava2020using,shrivastava2022a}. The idea is to fix the learned model weights and tune the input tensor to match the model output. Essentially, we optimize for the following loss
     $\argmin_{\hat{X}}\frac{1}{D}\norm{\hat{X}-f_\mathcal{N}(\hat{X})}_2^2 $
where, the input learnable tensor $\hat{X}$ is initialize randomly, for instance, from a Gaussian distribution $\sim \mathcal{N}(0, I)$. It is a parallel approach to \kals sampling, which is based on learning the input tensor via gradient descent while the model is frozen. It seemed to work well on an Infant Mortality data~\cite{CDC:InfantLinkedDatasets} for Neural Graphical Models~\cite{shrivastava2023neural} and their follow up work on Federated learning~\cite{chajewska2023federated}. We also want to underscore that the \textit{choice of architecture} plays an important role in the quality of samples generated. We tried these gradient-based sampling approaches for image generation using Multilayer Perceptron, Convolutional Neural Networks \& Transformer based architectures. In general, we found that one  will not always obtain a sample that looks similar to the input training data even though loss value goes down considerably. In other words, while running the sampling optimization, the input learnable tensor and model output matches, but the learned tensor does not visually match the training images (sometimes looks like noise). This suggests that the manifold learned by fitting the model to the input data points is too vast. Instead of randomly initializing $\hat{X}$, if we add some white noise to a data point from the training data, the sampling technique suggested, seems to work better. 

\textit{Exploring connection to Fractals theory}: Fractals are obtained by repeated applications of a rule or a function, given a starting state~\cite{hutchinson1981fractals}. It is indeed fascinating to read about the Mandelbrot set and the Julia set theory. We are, in a way, reverse engineering the underlying function, as repeated applications of it over an input noise eventually generates a sample. Another, very fascinating angle will be to investigate generation of images based on their `Hausdorff dimension'. We are currently investigating whether generating samples of lower fractal dimension is simpler. 


\section{Conclusions and Future Work}
We observed an `over-generalization' phenomenon in neural networks. We utilized it to introduce a new paradigm of generative modeling by creating a dataset  specific `Kaleidoscope'. Akin to fractal generation, we start with an input noise and our \kals sampling procedure generates samples by doing recursive functional calls. At the current stage, without any post-processing, we emphasize that our sampling results do not exceed the state-of-the-art Generative AI models. This can change, if we can better understand the manifold profiles for more specialized architectures used for image/audio/video analysis. We are exploring theoretical explanations, experiments on multimodal data, as well as conditional generation using the Generative Kaleidoscopic Networks.

\textit{Applications in video processing}: Video-based tasks such as action recognition~\cite{saini2022recognizing}, video segmentation~\cite{zhou2022survey}, object detection~\cite{jiao2021new}, identifying activity within videos~\cite{bodla2021hierarchical} and others are of great interest to the research community and have a considerable commercial impact~\cite{oprea2020review}. A natural application of the over-generalization phenomenon in conjunction with \kals sampling seems to be in the video representation and generation domain. Videos can be seen as a sequential collection of frames that are closely related. Our initial experimentation shows that the frame embeddings of a given video lie very close to each other in the manifold space. We are currently investigating Generative Kaleidoscopic Network's compatibility with the video prior representation frameworks provided in~\cite{shrivastava2024video2,shrivastava2024video3}. The \kals sampling procedure can be thought of as a stochastic process (or a temporal process). This idea can be extended to represent continuous multi-dimensional processes as done for videos in~\cite{shrivastava2024video1}. Furthermore, we are investigating diverse video generation tasks and the techniques provided in the following works~\cite{denton2018stochastic,shrivastava2021diverse,shrivastava2021diversethesis} seem to be good initial candidates.

\textit{Thanking note: I take this opportunity to give a hearty thanks a fellow computer science enthusiast Gaurav Shrivastava for our intriguing discussions about neural network properties.}

\textit{From the author's corner: There  exists a vast theoretical work for understanding the function signatures of neural networks and various deep learning architectures, which is also a subject of study in this paper. It is highly likely that I missed some important relevant works or similar analysis. Anyway, I had fun exploring neural network properties and I look forward to the research community's help in making this work more thorough in its future editions. Kindly reach out. Thanks!}

\bibliography{bibfile}
\bibliographystyle{iclr2023_conference}

\end{document}